\documentclass[10pt,twocolumn,letterpaper]{article}

\usepackage{cvpr}              

\usepackage{makecell} 

\usepackage[table]{xcolor}
\usepackage{times}
\usepackage{latexsym}
\usepackage{multirow}
\usepackage{colortbl}
\usepackage[T1]{fontenc}

\usepackage[utf8]{inputenc}
\usepackage{tabularx}
\usepackage{microtype}

\usepackage{inconsolata}

\usepackage{graphicx}
\usepackage{booktabs}
\usepackage{adjustbox}
\usepackage{amssymb}
\usepackage{microtype}
\usepackage{bbding}
\usepackage{algorithm}
\usepackage{algorithmic}

\definecolor{cvprblue}{rgb}{0.21,0.49,0.74}
\usepackage[pagebackref,breaklinks,colorlinks,allcolors=cvprblue]{hyperref}

\title{Accelerating Streaming Video Large Language Models  via \\ Hierarchical Token Compression}

\author{Yiyu Wang$^{1}$\thanks{Equal contribution. $\dagger$ Project leader. \text{\Envelope} Corresponding author} \quad Xuyang Liu$^{1,2*,\dagger}$ \quad Xiyan Gui$^{1,3}$ \quad Xinying Lin$^{4}$ \quad Boxue Yang$^{1}$ \\ \quad Chenfei Liao$^{1,5}$ \quad Tailai Chen$^{1}$ \quad Linfeng Zhang$^{1}$$^{\text{\Envelope}}$ \\
$^{1}$EPIC Lab, Shanghai Jiao Tong University \quad
$^{2}$Sichuan University \quad \\
$^{3}$Huazhong University of Science and Technology \quad $^{4}$Sun Yat-sen University \quad \\ $^{5}$Hong Kong University of Science and Technology (Guangzhou) \\ 
\textbf{Code:} \url{https://github.com/lern-to-write/STC}
}

\begin{document}
\maketitle

\begin{abstract}

Streaming Video Large Language Models (VideoLLMs) have demonstrated impressive performance across various video understanding tasks, but they face significant challenges in real-time deployment due to the high computational cost of processing dense visual tokens from continuous video streams. In streaming video scenarios, the primary bottleneck lies in the Vision Transformer (ViT) encoding stage, where redundant processing of temporally similar frames leads to inefficiency. Additionally, inflated token sequences during LLM pre-filling further exacerbate latency and memory overhead. To address these challenges, we propose \textbf{S}treaming \textbf{T}oken \textbf{C}ompression (\textbf{STC}), a plug-and-play hierarchical framework that seamlessly integrates into existing streaming VideoLLMs, optimizing both ViT encoding and LLM pre-filling stages to accelerate processing. STC introduces two token-level accelerators: \textbf{STC-Cacher}, which reduces ViT encoding overhead by caching and reusing features from temporally similar frames, and \textbf{STC-Pruner}, which compresses the visual token sequence before it enters the LLM, preserving only the most salient tokens based on both spatial and temporal relevance. Extensive experiments on four baseline streaming VideoLLMs across five benchmarks demonstrate that STC outperforms other compression methods. Notably, STC retains up to \textbf{99\%} of accuracy on the ReKV framework while reducing ViT encoding latency and LLM pre-filling latency by \textbf{24.5\%} and \textbf{45.3\%}.

\end{abstract}
\section{Introduction}

Video large language models (VideoLLMs)~\cite{zhang2025videollama,wang2025internvideo2.5,zhang2024llava-video,chen2024longvila,wang2025videoitg} have demonstrated strong performance across diverse video understanding tasks~\cite{li2024mvbench,fu2025videomme,zhou2024mlvu,wu2024longvideobench}. 
Recently, emerging applications such as live sports commentary~\cite{rao2025soccer} and augmented reality glasses~\cite{wen2025glasses} have created an urgent demand for streaming video understanding (SVU), where models must process incoming video frames continuously and generate responses with minimal latency~\cite{chen2024videollm,qian2025dispider}. 
However, the computational cost of processing dense visual tokens from multi-frame inputs renders existing VideoLLMs too slow for real-time streaming scenarios, severely limiting their deployment in latency-sensitive applications~\cite{yao2025timechat}.

\begin{figure}[t]
    \centering
    \includegraphics[width=\linewidth]{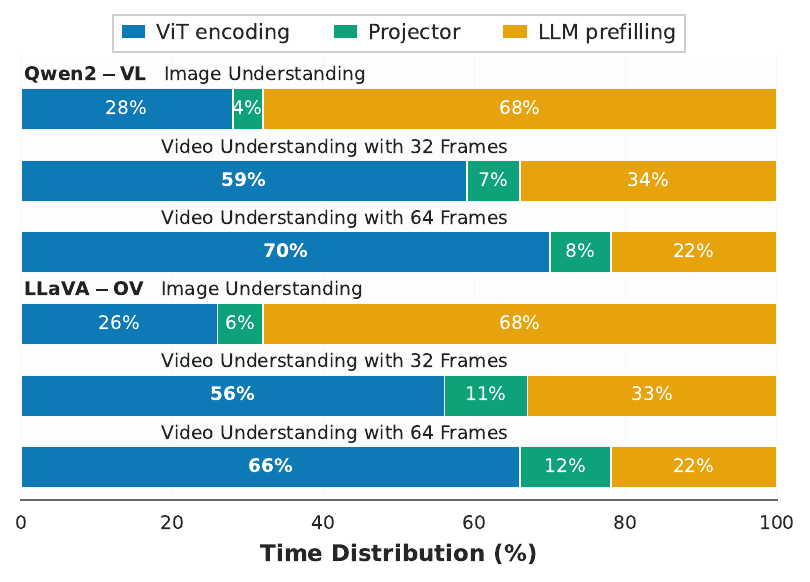}
    \vspace{-7mm}
    \caption{\textbf{Inference time breakdown across components in various vision-language understanding scenarios.} ViT encoding typically accounts for a substantial fraction of the inference time in video understanding, about \textbf{2-3 times} that in image understanding.}
    \vspace{-5mm}
    \label{fig:motivation}
\end{figure}

To address this, two main approaches have been explored. Token compression methods reduce visual tokens either before~\cite{tao2025dycoke,Yang2024:Visionzip,liu2025vidcom2} or within the LLM~\cite{xing2024pdrop,wen2025dart,chen2025variation}, improving both prefill and decoding efficiency. Key-Value (KV) cache compression methods~\cite{feng2024adakv,liu2025mixkv,tao2025plug} evict less important KV pairs to reduce memory overhead during decoding. However, in streaming scenarios where frames arrive continuously, the computational bottleneck lies primarily in the vision encoder (ViT~\cite{dosovitskiy2020vit}). Since each incoming frame must be encoded independently through ViT, the repeated ViT inference dominates overall latency before tokens even reach the LLM.

Figure~\ref{fig:motivation} demonstrates this phenomenon: the inference time breakdown across components reveals a striking difference between image and video understanding. For models such as Qwen2-VL~\cite{Wang:Qwen2-VL} and LLaVA-OV~\cite{li2024llava-ov}, ViT encoding in video understanding incurs \textbf{2-3 times} higher computational cost compared to image understanding, making it the dominant latency contributor. Moreover, when LLaVA-OV processes a 32-frame video, this results in $32 \times 196 = 6{,}272$ visual tokens fed into the LLM during prefilling. In comparison, image understanding tasks such as MMBench~\cite{Liu:MMBench} typically require the LLM to process only around 1{,}900 visual tokens. The over \textbf{three-fold} increase translates directly to substantially longer latency, presenting a critical barrier for real-time streaming deployment.
These observations underscore that \textit{compression methods must reduce both ViT encoding overhead and context sequence length to enable efficient streaming video understanding}. To better understand the unique characteristics of streaming scenarios and guide our method design, we conduct an empirical analysis comparing streaming and offline video understanding, revealing \textbf{\textit{two key features}} in streaming videos:

\begin{figure}[t]
    \centering
    \includegraphics[width=\linewidth]{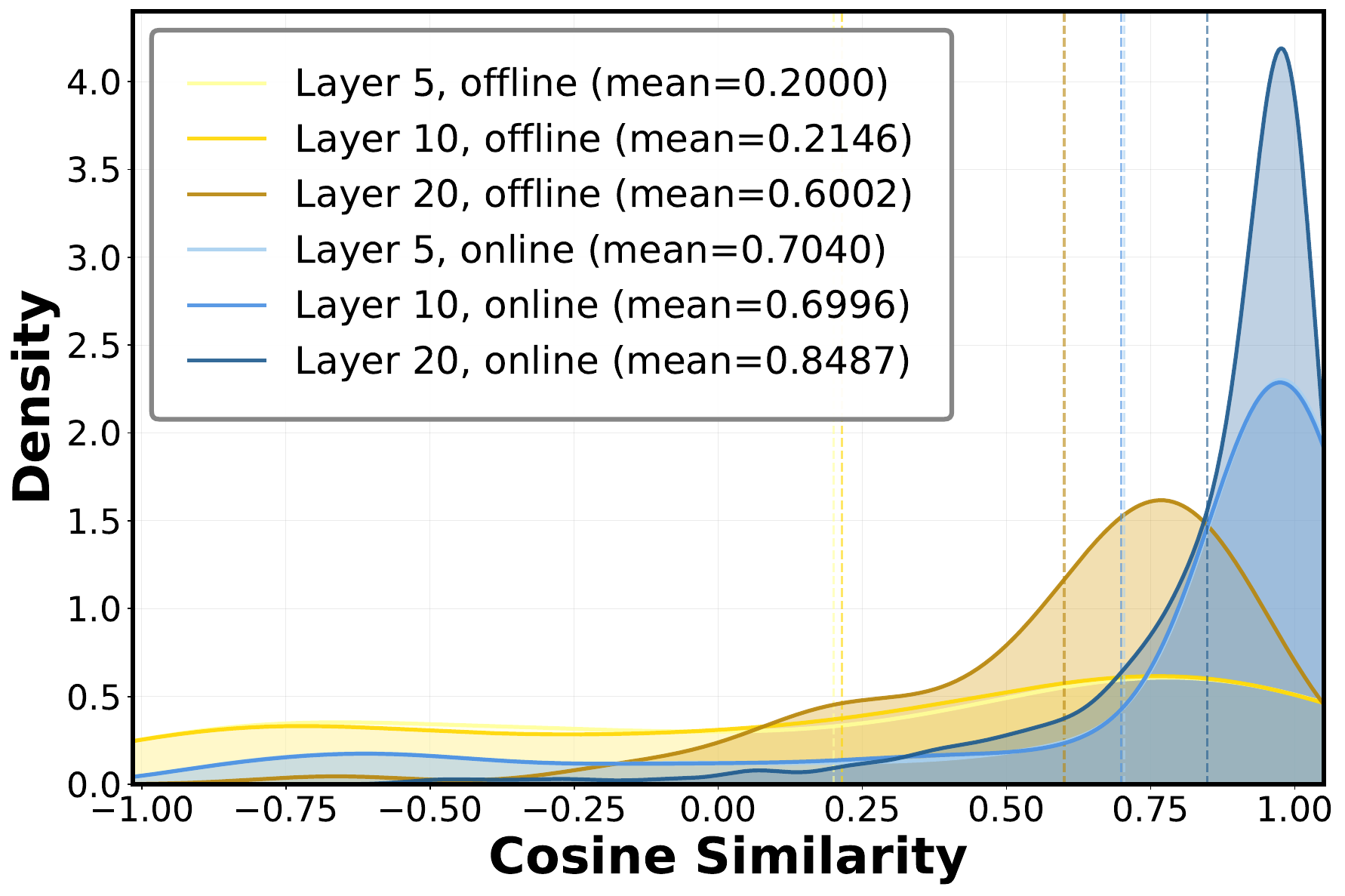}
    \vspace{-7mm}
    \caption{\textbf{Temporal redundancy in adjacent frames in ViT encoding.} Streaming videos (``online'') tend to show higher similarity than offline videos, indicating higher temporal redundancy.}
    \vspace{-5mm}
    \label{fig:redundancy}
\end{figure}

\noindent \textbf{(I) Temporal Redundancy in ViT Encoding:} Streaming videos require denser frame sampling, leading to substantial temporal redundancy. Figure~\ref{fig:redundancy} compares streaming (0.5fps) and offline videos (64 frames) during ViT encoding by LLaVA-OV~\cite{li2024llava-ov}. At deeper layers (\textit{e.g.}, Layer 20), the mean cosine similarity of adjacent frame features in streaming reaches 0.85, compared to 0.60 in offline settings. This difference arises because \textit{consecutive frames in streaming often capture \textbf{nearly identical} visual content}. However, existing methods fail to exploit this redundancy, as most ignore it during ViT encoding and operate only on final token representations. A few methods, such as ToMe~\cite{Bolya:ToMe}, perform token merging within ViT layers, but this disrupts encoding, leading to performance degradation for SVU (see Table~\ref{tab:ovo-bench},~\ref{tab:streamingbench}).

\noindent \textbf{(II) Incomplete Video and Unknown Instructions:} Streaming scenarios impose a constraint: \textit{models cannot access complete video content in advance, nor can they know user instructions beforehand}. This makes existing compression strategies ineffective. Methods relying on global video features for token selection~\cite{liu2025vidcom2,shao2025holitom,ma2025mmg} cannot operate efficiently, as they require the entire video for compression decisions. Instruction-aware approaches~\cite{Chen:FastV,xing2024pdrop,Zhang:SparseVLM}, which prune tokens based on query relevance, fail in streaming, where queries arrive only after frames are processed~\cite{yang2025streammem}. These challenges highlight a gap: streaming video understanding demands compression methods that operate \textbf{causally}, exploiting temporal redundancy during encoding while adapting to incrementally arriving frames, without relying on global context or future instructions.

Based on the above analysis, we propose \textbf{S}treaming \textbf{T}oken \textbf{C}ompression (\textbf{STC}), a plug-and-play hierarchical acceleration framework that jointly optimizes the ViT encoding and LLM prefilling stages for efficient SVU. STC consists of two \textit{orthogonal} modules. First, \textbf{STC-Cacher} reduces ViT forward passes by identifying temporally redundant frames through spatial feature similarity. It caches key frame visual features and reuses them for similar frames, avoiding redundant computations. Second, \textbf{STC-Pruner} compresses the visual token sequence after the vision encoder, preserving tokens salient in both spatial and temporal dimensions, and pruning redundant ones to shorten the LLM prefill sequence. Together, these modules enable low-latency video understanding while maintaining high semantic fidelity. Our contributions are \textbf{four-fold}:

\begin{itemize}
    \item \textbf{Empirical Analysis of Streaming Scenarios:} We empirically analyze limitations of existing compression methods for SVU, revealing the need for causal compression that reduces ViT cost without relying on future instructions.
    
    \item \textbf{Plug-and-Play Acceleration Framework:} We propose Streaming Token Compression (STC), a plug-and-play framework that seamlessly integrates into existing VideoLLMs for efficient streaming video understanding.
    
    \item \textbf{Two Complementary Token Compressors:} STC introduces two modules: STC-Cacher accelerates ViT encoding by caching features from adjacent frames, and STC-Pruner accelerates LLM prefill by pruning low-saliency tokens.
    
    \item \textbf{Comprehensive Validation and Results:} Extensive experiments demonstrate STC achieves superior performance-efficiency trade-offs, retaining 99\% accuracy on ReKV while reducing LLM prefill latency by 45.3\%.
\end{itemize}

\section{Related Work}

\subsection{Streaming Video Understanding}

Streaming Video Understanding (SVU) requires VideoLLMs~\cite{li2024llava-ov, zhang2024llava-video, bai2025qwen2.5-vl, wang2025videoitg, zhang2025videollama} to continuously process incoming video content and respond to user queries in real-time~\cite{di2025streaming, chen2025livecc, niu2025ovo, liu2024streamchat}. SVU systems can generally be categorized into two main paradigms: \textbf{(i)} \textit{End-to-end online models} trained specifically for streaming, such as Dispider~\cite{qian2025dispider}, which decouples perception and decision-making, and StreamForest~\cite{zeng2025streamforest}, which utilizes a Persistent Event Memory Forest for long-term memory retention and a Fine-grained Spatiotemporal Window for real-time comprehension. LiveCC~\cite{chen2025livecc} combines streaming ASR transcripts and video frames for continuous training and real-time commentary. While these models are effective in many contexts, they come with high training costs and offer limited compatibility with offline VideoLLMs, making them less efficient for broader use cases. \textbf{(ii)} \textit{Offline-to-online frameworks} adapt pre-trained VideoLLMs~\cite{li2024llava-ov, Wang:Qwen2-VL, bai2025qwen2.5-vl} to streaming environments without the need for retraining~\cite{yang2025streammem, ning2025livevlm, yang2025streamagent}. ReKV~\cite{di2025streaming} pioneers this approach by maintaining frame-wise KV caches and employing sliding-window encoding for real-time video question answering. LiveVLM~\cite{ning2025livevlm} and StreamChat~\cite{liu2024streamchat} focus on preserving temporal coherence through effective memory management, enhancing the overall streaming experience.

Despite their success, these approaches still process each frame through ViT encoding, feeding dense token sequences into the LLM, resulting in high computational and memory costs that limit real-time deployment.

\subsection{Token Compression for VideoLLMs}
Token compression has emerged as a promising strategy to accelerate VideoLLM inference by reducing redundant visual tokens~\cite{Yang2024:Visionzip, Zhang:SparseVLM, Chen:FastV, liu2025globalcom2, tao2025dycoke,shao2025holitom,shao2025tokens,liu2025shifting}. Existing methods can be broadly categorized into two approaches for VideoLLMs: \textbf{(i)} \textit{Offline compression}, which assumes full access to the entire video. Methods like ToMe~\cite{bolya2022token} merge similar tokens during ViT encoding, while post-encoding methods select salient tokens after feature extraction~\cite{Chen:FastV, Yang2024:Visionzip, han2024ficoco}. The former accelerates both ViT encoding and LLM prefiling, while the latter only accelerates prefiling, leaving ViT encoding inefficient. More critically, methods like SparseVLM~\cite{Zhang:SparseVLM}, DyCoke~\cite{tao2025dycoke}, and VidCom$^2$~\cite{liu2025video} either rely on global visibility of the entire video or require user input, making them incompatible with causal streaming constraints.
\textbf{(ii)} \textit{Streaming compression} adapts compression for streaming settings but remains limited. LiveVLM~\cite{ning2025livevlm} compresses the KV cache to reduce memory usage but still cannot address the high cost of ViT encoding and long sequence lengths. TimeChat-Online~\cite{yao2025timechat} drops tokens based on adjacent-frame similarity, but this short-horizon strategy fails to capture long-range redundancy and often retains repetitive content.

In summary, prior works either overlook the causal nature of streaming or optimize only one stage. Our work is \textbf{the first} to jointly optimize ViT encoding and LLM prefill with a streaming-native, two-stage design that respects temporal causality while maximizing redundancy removal.

\section{Methodology}

\begin{figure}[t]
    \centering
    \includegraphics[width=\linewidth]{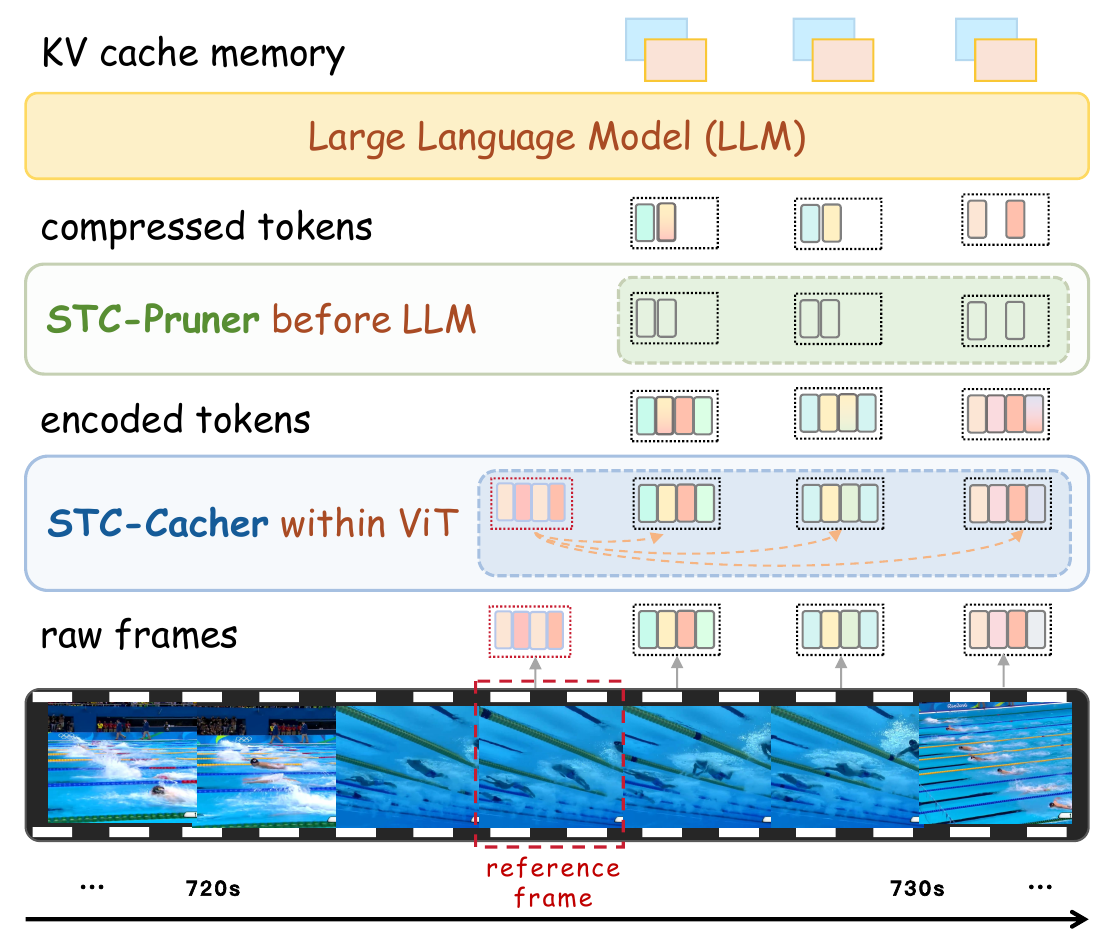}
    \vspace{-6mm}
    \caption{\textbf{Overview of Streaming Token Compression (STC).} Our framework accelerates streaming Video-LLMs in two stages. \textbf{STC-Cacher} employs selective recomputation to reduce computational redundancy in the ViT. \textbf{STC-Pruner} then reduces the token sequence to alleviate the prefilling latency for the LLM.}
    \vspace{-5mm}
\label{fig:overview}
\end{figure}

\subsection{Preliminary}

\noindent \textbf{Streaming Inference with VideoLLMs.}
Most VideoLLMs, initially designed for offline video understanding (OVU)~\cite{li2024llava-ov, zhang2024llava-video, zhang2025videollama, shu2024video-xl, qin2025video-xl-2}, can be adapted for SVU by processing video in discrete chunks. A continuous video stream $\mathbf{V} = \left\{ \mathbf{v}_t \right\}_{t=1}^T \in \mathbb{R}^{T \times H \times W \times 3}$ is divided into segments, each corresponding to a chunk $\mathbf{V}_c = \left\{ \mathbf{v}_t \right\}_{t=cL+1}^{(c+1)L}$, where $L$ is the chunk length. For each chunk, a vision encoder (e.g., ViT) extracts visual features as token embeddings $\mathbf{Z}_c = \left\{ \mathbf{z}_t \right\}_{t=cL+1}^{(c+1)L} \in \mathbb{R}^{L \times N \times D}$, where $N$ is the number of patches and $D$ is the embedding dimension.

These token embeddings $\mathbf{Z}_c$ are projected into the LLM’s embedding space for autoregressive processing. The LLM uses visual tokens $\mathbf{X}_c^v$ and any prior textual context $\mathbf{X}_c^t$ to predict subsequent tokens. To maintain temporal consistency under causal constraints, KV states generated from each chunk are cached in a memory bank $\mathcal{M}$, which serves as context for future processing. Specifically, the KV states $\mathbf{KV}_c = \left\{ \mathbf{K}_c, \mathbf{V}_c \right\}$, where $\mathbf{K}_c$ and $\mathbf{V}_c$ represent key and value embeddings, are stored and used to condition the model’s future generations.

\begin{figure*}[!t]
  \centering
   \includegraphics[width=\linewidth]{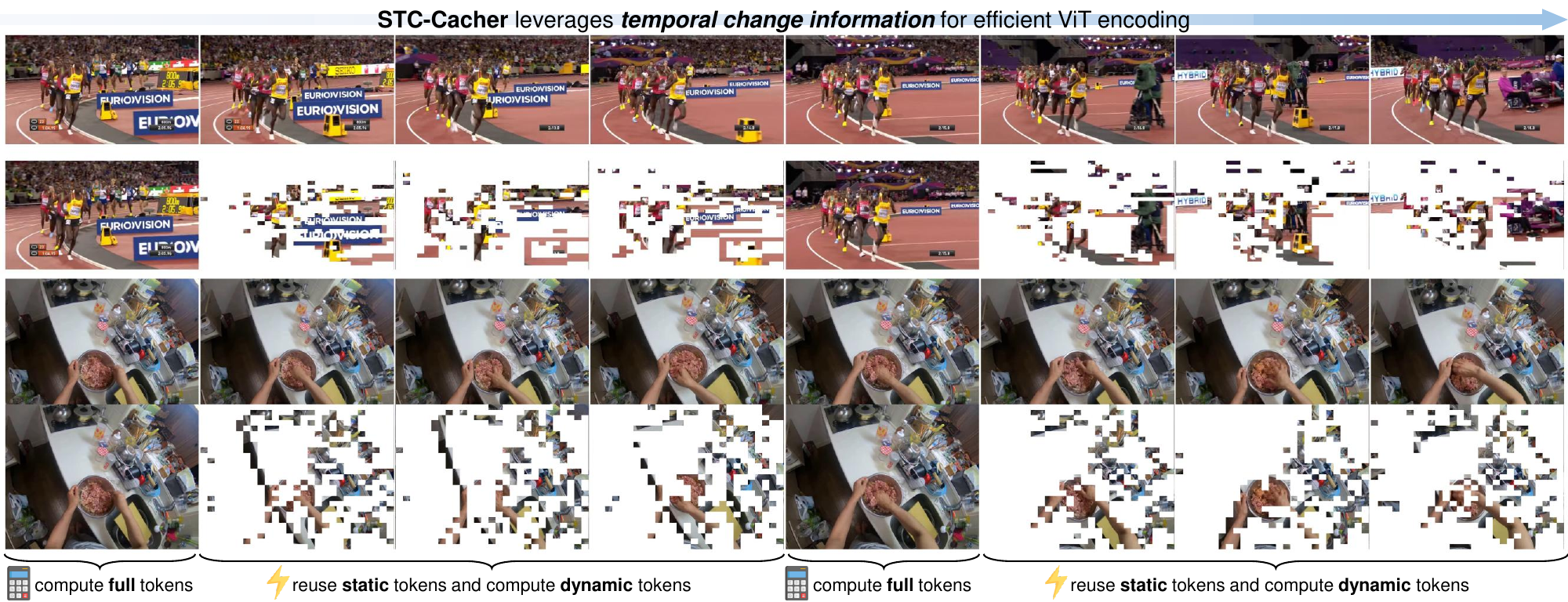}
   \vspace{-7mm}
    \caption{\textbf{Visualization of cache-aware selective computation by STC-Cacher.} For reference frames, STC-Cacher computes and caches all tokens. For subsequent frames, only dynamic tokens are computed, while static tokens reuse cached features from reference frames.}   
   \label{fig:caching_visualizations}
   \vspace{-5mm}
\end{figure*}

This approach enables continuous video processing while respecting causal constraints, supporting real-time video understanding in streaming settings.

\noindent \textbf{The Redundancy Bottleneck.}
This sequential processing paradigm introduces computational bottlenecks. We identify two main sources of redundancy:
\begin{itemize}
    \item \textbf{Temporal Redundancy in ViT Encoding:} In streaming video settings, adjacent frames often share similar visual content, such as static backgrounds, leading to redundant computations in ViT encoding. Processing static information provides diminishing returns in new insights.
    \item \textbf{Long Context Redundancy in LLM Prefilling:} As the video progresses, redundancy accumulates in the token sequence fed to the LLM. This includes persistent temporal features (\textit{e.g.}, background) and low-information regions within frames. This redundancy burdens the LLM prefilling stage, characterized by the quadratic computational complexity $\mathcal{O}(N^2)$ of self-attention~\cite{vaswani2017attention}, and inflates the memory footprint of the KV cache $\mathcal{M}$ without proportional gains in comprehension.
\end{itemize}
These observations motivate a unified compression strategy to reduce redundancy at both the vision encoding and language processing stages, while adhering to the causal constraints of streaming inference.

\subsection{Streaming Token Compression}

To address the redundancy bottlenecks discussed above, we propose \textbf{S}treaming \textbf{T}oken \textbf{C}ompression (STC), a plug-and-play acceleration framework for efficient SVU. As shown in Figure~\ref{fig:overview}, STC tackles redundancy through two complementary token-level compression components:

\begin{itemize}
    \item \textbf{STC-Cacher} combines \textit{temporal change information} to \textbf{cache} and reuse visual features for static content, selectively computing only the temporally dynamic tokens, thus addressing temporal redundancy in ViT encoding.
    \item \textbf{STC-Pruner} combines \textit{causal event information} to \textbf{prune} redundant tokens in the visual token sequence before it enters the LLM, ensuring precise token sequence compression to reduce computational cost.
\end{itemize}

Both components are \textbf{query- and future-agnostic}, ensuring compatibility with real-time streaming constraints. Together, they reduce computational overhead while preserving causal integrity for SVU. Below, we elaborate on the detailed operations of these two components.

\subsection{STC-Cacher: Selective Computation in ViT}

\begin{figure}[t]
    \centering
    \includegraphics[width=\linewidth]{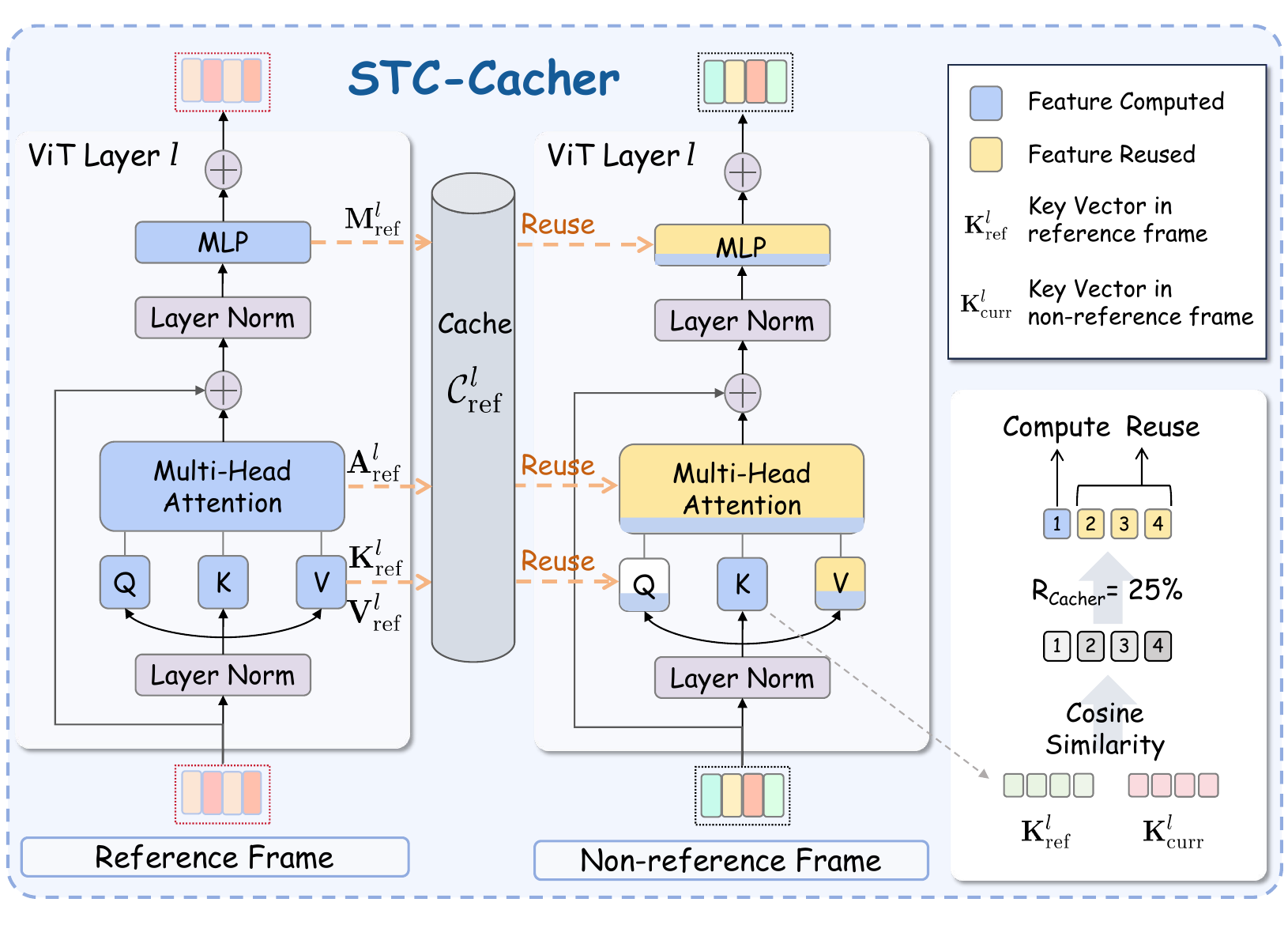}
    \vspace{-7mm}
    \caption{\textbf{The Mechanism of STC-Cacher.} Instead of a full forward pass, STC-Cacher identifies novel tokens by comparing their Key projections ($K_{\text{curr}}$) to a cached reference ($K_{\text{ref}}$). It then selectively recomputes only the Query and Value representations for these dynamic tokens and \textbf{scatters} Value into the cached Value matrix for an efficient, low-rank update attention mechanism.}
    \vspace{-5mm}
\label{fig:STC-cacher}
\end{figure}

The computational cost of ViT encoding in streaming scenarios is primarily driven by the repetitive processing of temporally redundant frames. While naive strategies like token pruning or merging~\cite{Liang:EViT, Bolya:ToMe} reduce computational cost, they result in significant information loss during the visual encoding process. This leads to our first research question:

\vspace{0.3em}
\textit{\textbf{How can we reduce the computational cost of ViT encoding on sequential frames while preserving temporal information content from the original video stream?}.}
\vspace{0.3em}

To address this, we introduce \textbf{STC-Cacher} (Figure~\ref{fig:STC-cacher}), a \textbf{cache-aware selective computation} strategy designed to focus computation on dynamic tokens while reusing cached states for static content. The process is detailed below:

\noindent \textbf{(I) Reference Frame (Full Computation and Caching):}  
For a reference frame $\mathbf{f}_{\text{ref}}$, we perform a full forward pass, caching intermediate representations at each ViT layer to serve as a reference for subsequent frames. The cached information $\mathcal{C}_{\text{ref}}^l$ at layer $l$ includes:
\begin{equation}
\mathcal{C}_{\text{ref}}^l = \{ \mathbf{K}_{\text{ref}}^l, \mathbf{V}_{\text{ref}}^l, \mathbf{A}_{\text{ref}}^l, \mathbf{M}_{\text{ref}}^l \}
\end{equation}
These representations are used as reference when processing non-reference frames.

\noindent \textbf{(II) Non-reference Frame (Selective Computation):}  
For subsequent frames $\mathbf{F}_{\text{new}}$, we leverage the cached information $\mathcal{C}_{\text{ref}}^l$ to bypass most computations. Instead of a full forward pass, we focus on the \textbf{dynamic tokens}, which contain novel information compared to the cached reference.

We introduce two hyper-parameters: the \textit{cache interval} $\mathcal{N}$ and the \textit{cache reuse ratio} $R_\text{Cacher}$. The cache interval $\mathcal{N}$ determines how frequently the reference frame is updated, and $R_\text{Cacher}$ controls the proportion of dynamic tokens selected. For instance, with $\mathcal{N} = 4$ and $R_\text{Cacher} = 75\%$, every 4th frame is chosen as a reference, and 25\% of tokens are computed as dynamic, while the remaining 75\% are reused from the cache. The process is as follows:

\noindent \textbf{(i) Identify Dynamic Tokens:}  
We compute the cosine similarity between the current Key projections $\mathbf{K}_{\text{curr}, \mathbf{f}}^l$ and the cached reference $\mathbf{K}_{\text{ref}}^l$:
\begin{equation}
S_{\mathbf{f}} = \frac{\mathbf{K}_{\text{curr}, \mathbf{f}}^l \cdot \mathbf{K}_{\text{ref}}^l}{\|\mathbf{K}_{\text{curr}, \mathbf{f}}^l\| \|\mathbf{K}_{\text{ref}}^l\|} \in \mathbb{R}^{T}
\end{equation}
We select the top $k$ tokens with the lowest similarity scores as the dynamic set:
\begin{equation}
\mathcal{I}_{\mathbf{f}} = \underset{i \in \{1...T\}}{\text{arg top-k}} \left( 1 - S_{\mathbf{f}}[i] \right)
\end{equation}
where $k = \lfloor T \cdot r \rfloor$ and $r$ is determined by $R_\text{Cacher}$.

\begin{figure}[t]
    \centering
    \includegraphics[width=\linewidth]{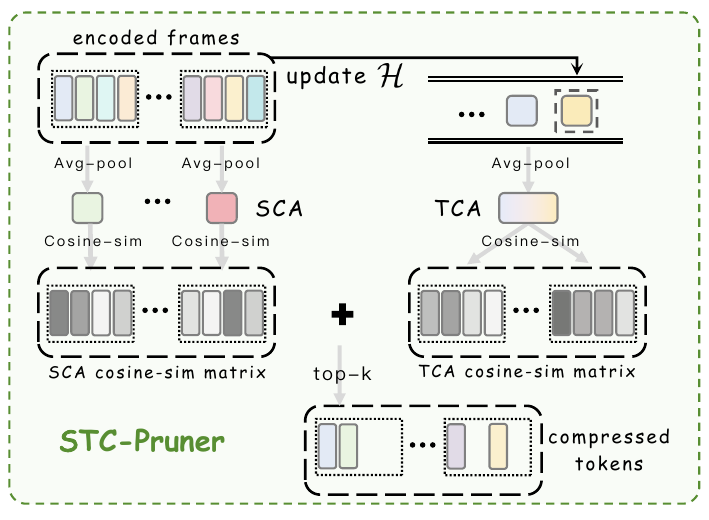}
    \vspace{-7mm}
    \caption{\textbf{The Mechanism of STC-Pruner.} To accelerate LLM prefilling, STC-Pruner scores each token based on its novelty. Novelty is measured as the joint dissimilarity to a \textbf{Temporal Context Anchor} (TCA), representing historical context, and a \textbf{Spatial Context Anchor} (SCA), representing the current frame's global context. Only tokens with high novelty scores are retained.}
    \vspace{-5mm}
\label{fig:STC-Pruner}
\end{figure}

\noindent \textbf{(ii) Selective Attention:}  
We compute the Query and Key only for dynamic tokens indexed by $\mathcal{I}_{\mathbf{f}}$:
\begin{equation}
Q_{\text{sel}, \mathbf{f}}^l = Q(\text{LN}_1(\mathbf{X}_{\mathbf{f}}^l))[\mathcal{I}_{\mathbf{f}}], 
V_{\text{sel}, \mathbf{f}}^l = V(\text{LN}_1(\mathbf{X}_{\mathbf{f}}^l))[\mathcal{I}_{\mathbf{f}}]
\end{equation}
We construct the full Key by initializing it from the cached reference and \textbf{scattering} the newly computed Key values:
\begin{equation}
\bar{V}_{\mathbf{f}}^l \leftarrow \mathbf{V}_{\text{ref}}^l; \quad \bar{V}_{\mathbf{f}}^l[\mathcal{I}_{\mathbf{f}}] \leftarrow V_{\text{sel}, \mathbf{f}}^l
\end{equation}
Attention is then computed using the selective queries $Q_{\text{sel}, \mathbf{f}}^l$ and the full Key matrix $\bar{V}_{\mathbf{f}}^l$:
\begin{equation}
A_{\text{sel}, \mathbf{f}}^l = \text{Attention}(Q_{\text{sel}, \mathbf{f}}^l, \bar{K}_{\mathbf{f}}^l, \mathbf{V}_{\text{curr}, \mathbf{f}}^l)
\end{equation}

\noindent \textbf{(iii) Scatter-Update Output:}  
The attention output is updated by scattering the results into the cached attention $\mathbf{A}_{\text{ref}}^l$:
\begin{equation}
\bar{A}_{\mathbf{f}}^l \leftarrow \mathbf{A}_{\text{ref}}^l; \quad \bar{A}_{\mathbf{f}}^l[\mathcal{I}_{\mathbf{f}}] \leftarrow A_{\text{sel}, \mathbf{f}}^l
\end{equation}
This updated tensor is passed to the MLP block, where a similar process is applied.

As depicted in Figure~\ref{fig:caching_visualizations}, by focusing computation on dynamic tokens and reusing cached values for static content, STC-Cacher significantly reduces the computational load of ViT encoding in streaming scenarios. In this way, STC-Cacher preserves the temporal information content while maintaining efficient processing, leading to reduced encoding time and memory usage.

\definecolor{greenrightcolor}{RGB}{0,144,81}
\definecolor{redrightcolor}{RGB}{255,0,0}
\definecolor{bluerightcolor}{RGB}{0,0,255}

\newcommand{\downtinya}[1]{\!\textcolor{greenrightcolor}{\tiny #1}}
\newcommand{\uptiny}[1]{\!\textcolor{redrightcolor}{\tiny #1}}
\newcommand{\downtinyb}[1]{\!\textcolor{bluerightcolor}{\tiny #1}}

\begin{table*}[t]
\small
\centering
\setlength{\abovecaptionskip}{0.2cm}
\setlength{\tabcolsep}{3pt}
\begin{adjustbox}{max width=2.1\columnwidth}

\begin{tabular}{l|ccccccc|cccc|cccc|c|cc}

\toprule

\multirow{2}{*}{\textbf{Methods}} &
\multicolumn{7}{c|}{\textbf{Real-Time Visual Perception}} &
\multicolumn{4}{c|}{\textbf{Backward Tracing}} &
\multicolumn{4}{c|}{\textbf{Forward Active Respond.}} &  
\multirow{2}{*}{\textbf{Overall}} &  
\multirow{2}{*}{\textbf{\shortstack{ViT Enc. \\ Latency}}} &  
\multirow{2}{*}{\textbf{\shortstack{LLM Pref. \\ Latency}}} \\

& OCR & ACR & ATR & STU & FPD & OJR & Avg. &
EPM & ASI & HLD & Avg. &
REC & SSR & CRR & Avg. &   
& & \\

\midrule

\multicolumn{19}{c}{\textbf{End-to-End Online VideoLLMs with STC-Cacher}} \\

\midrule

\textcolor[rgb]{ .502,  .502,  .502}{Dispider\tiny\texttt{(CVPR25)}} & \textcolor[rgb]{ .502,  .502,  .502}{49.7} & \textcolor[rgb]{ .502,  .502,  .502}{44.0} & \textcolor[rgb]{ .502,  .502,  .502}{60.3} & \textcolor[rgb]{ .502,  .502,  .502}{43.8} & \textcolor[rgb]{ .502,  .502,  .502}{59.4} & \textcolor[rgb]{ .502,  .502,  .502}{48.4} & \textcolor[rgb]{ .502,  .502,  .502}{51.0} & \textcolor[rgb]{ .502,  .502,  .502}{47.8} & \textcolor[rgb]{ .502,  .502,  .502}{54.7} & \textcolor[rgb]{ .502,  .502,  .502}{5.4} & \textcolor[rgb]{ .502,  .502,  .502}{36.0} & \textcolor[rgb]{ .502,  .502,  .502}{16.9} & \textcolor[rgb]{ .502,  .502,  .502}{40.1} & \textcolor[rgb]{ .502,  .502,  .502}{45.8} & \textcolor[rgb]{ .502,  .502,  .502}{34.3} & \textcolor[rgb]{ .502,  .502,  .502}{40.4} &
\textcolor[rgb]{ .502,  .502,  .502}{26.4} &
\textcolor[rgb]{ .502,  .502,  .502}{115.9}\\

\rowcolor[HTML]{DAEFF9}
\textbf{+STC-Cacher} & 50.7 & 39.4 & 55.9 & 41.7 & 58.4 & 48.5 & 49.1 & 47.5 & 53.7 & 5.4 & 35.2 & 17.2 & 36.6 & 45.3 & 33.4 & 39.2 & 18.9\downtinya{(↓28.4\%)} & 115.9 \\

\midrule

\textcolor[rgb]{ .502,  .502,  .502}{LiveCC\tiny\texttt{(CVPR25)}} & \textcolor[rgb]{ .502,  .502,  .502}{68.5} & \textcolor[rgb]{ .502,  .502,  .502}{39.4} & \textcolor[rgb]{ .502,  .502,  .502}{62.1} & \textcolor[rgb]{ .502,  .502,  .502}{43.8} & \textcolor[rgb]{ .502,  .502,  .502}{68.3} & \textcolor[rgb]{ .502,  .502,  .502}{59.8} & \textcolor[rgb]{ .502,  .502,  .502}{57.0} & \textcolor[rgb]{ .502,  .502,  .502}{58.2} & \textcolor[rgb]{ .502,  .502,  .502}{59.5} & \textcolor[rgb]{ .502,  .502,  .502}{89.2} & \textcolor[rgb]{ .502,  .502,  .502}{69.0} & \textcolor[rgb]{ .502,  .502,  .502}{30.9} & \textcolor[rgb]{ .502,  .502,  .502}{56.4} & \textcolor[rgb]{ .502,  .502,  .502}{72.1} & \textcolor[rgb]{ .502,  .502,  .502}{53.2} & \textcolor[rgb]{ .502,  .502,  .502}{59.7} &
\textcolor[rgb]{ .502,  .502,  .502}{181.2}&
\textcolor[rgb]{ .502,  .502,  .502}{818.4}\\

\rowcolor[HTML]{DAEFF9}
\textbf{+STC-Cacher} & 61.7 & 36.7 & 59.5 & 43.8 & 67.3 & 53.8 & 53.8 & 56.6 & 56.1 & 87.6 & 66.8 & 29.8 & 54.2 & 70.0 & 51.3 & 57.3 & 126.84\downtinya{(↓30.0\%)} & 818.4 \\

\midrule

\textcolor[rgb]{ .502,  .502,  .502}{StreamForest\tiny\texttt{(NIPS25)}} & \textcolor[rgb]{ .502,  .502,  .502}{71.8} & \textcolor[rgb]{ .502,  .502,  .502}{51.4} & \textcolor[rgb]{ .502,  .502,  .502}{72.4} & \textcolor[rgb]{ .502,  .502,  .502}{46.6} & \textcolor[rgb]{ .502,  .502,  .502}{67.3} & \textcolor[rgb]{ .502,  .502,  .502}{59.8} & \textcolor[rgb]{ .502,  .502,  .502}{61.6} & \textcolor[rgb]{ .502,  .502,  .502}{59.3} & \textcolor[rgb]{ .502,  .502,  .502}{63.5} & \textcolor[rgb]{ .502,  .502,  .502}{33.3} & \textcolor[rgb]{ .502,  .502,  .502}{52.0} & \textcolor[rgb]{ .502,  .502,  .502}{32.5} & \textcolor[rgb]{ .502,  .502,  .502}{70.8} & \textcolor[rgb]{ .502,  .502,  .502}{56.7} & \textcolor[rgb]{ .502,  .502,  .502}{53.3} & \textcolor[rgb]{ .502,  .502,  .502}{54.3} &
\textcolor[rgb]{ .502,  .502,  .502}{103.7} &
\textcolor[rgb]{ .502,  .502,  .502}{366.2} \\

\rowcolor[HTML]{DAEFF9}
\textbf{+STC-Cacher} & 66.4 & 49.5 & 65.5 & 46.1 & 66.3 & 60.9 & 59.1 & 57.9 & 62.2 & 32.8 & 51.5 & 29.9 & 68.2 & 56.3 & 51.5 & 52.3 & 67.7\downtinya{(↓34.7\%)} & 366.2 \\

\midrule

\multicolumn{19}{c}{\textbf{Offline-to-Online Framework ReKV with Token Compression Methods}} \\

\midrule

\textcolor[rgb]{ .502,  .502,  .502}{ReKV\tiny\texttt{(ICLR25)}} & \textcolor[rgb]{ .502,  .502,  .502}{73.8} & \textcolor[rgb]{ .502,  .502,  .502}{56.0} & \textcolor[rgb]{ .502,  .502,  .502}{74.1} & \textcolor[rgb]{ .502,  .502,  .502}{51.7} & \textcolor[rgb]{ .502,  .502,  .502}{70.3} & \textcolor[rgb]{ .502,  .502,  .502}{60.3} & \textcolor[rgb]{ .502,  .502,  .502}{64.4} & \textcolor[rgb]{ .502,  .502,  .502}{54.2} & \textcolor[rgb]{ .502,  .502,  .502}{57.4} & \textcolor[rgb]{ .502,  .502,  .502}{28.5} & \textcolor[rgb]{ .502,  .502,  .502}{46.7} & \textcolor[rgb]{ .502,  .502,  .502}{25.4} & \textcolor[rgb]{ .502,  .502,  .502}{64.6} & \textcolor[rgb]{ .502,  .502,  .502}{53.3} & \textcolor[rgb]{ .502,  .502,  .502}{47.8} & \textcolor[rgb]{ .502,  .502,  .502}{52.6} &
\textcolor[rgb]{ .502,  .502,  .502}{103.7} &
\textcolor[rgb]{ .502,  .502,  .502}{482.4}\\
+ToMe\tiny\texttt{(ICLR23)} & 61.1 & 49.5 & 52.6 & 42.7 & 62.4 & 50.0 & 53.1 & 49.2 & 50.0 & 29.6 & 42.9 & 19.2 & \underline{60.7} & 50.0 & 43.3 & 46.4 & 70.5\downtinya{(↓32\%)} & 257.8\downtinya{(↓46.6\%)}\\

+VisionZip\tiny\texttt{(CVPR25)} & 49.0 & 49.5 & 64.7 & 44.4 & 64.4 & 50.5 & 53.8 & 47.1 & \underline{54.1} & 30.7 & 44.0 & 21.9 & 58.4 & \textbf{53.8} & 44.7 & 47.5 & 103.7 & 258.3\downtinya{(↓46.5\%)}\\

+VidCom$^2$\tiny\texttt{(EMNLP25)} & \underline{65.8} & \textbf{59.6} & \underline{69.0} & \underline{47.2} & 64.4 & \underline{56.5} & 60.4 & \underline{50.5} & 53.4 & \underline{32.8} & \textbf{45.6} & 25.8 & 59.0 & 50.8 & 45.2 & 50.4 & 103.7 & 259.1\downtinya{(↓46.3\%)}\\

\rowcolor[HTML]{DAEFF9}
\textbf{+STC-Pruner} & 64.4 & \textbf{59.6} & 68.1 & \textbf{48.9} & \underline{65.4} & \underline{56.5} & \underline{60.5} & \textbf{51.2} & 52.0 & \textbf{33.3} & \underline{45.5} & \underline{25.9} & 59.3 & 52.1 & \underline{45.8} & \underline{50.6} & 103.7 & 259.2\downtinya{(↓46.3\%)}\\

\rowcolor[HTML]{A8D7F4}
\textbf{+STC-Cacher \& Pruner} & \textbf{68.5} & \underline{57.8} & \textbf{73.3} & \underline{47.2} & \textbf{68.3} & \textbf{59.8} & \textbf{62.5} & \underline{50.5} & \textbf{55.4} & 30.1 & 45.3 & \textbf{27.9} & \textbf{63.3} & \underline{52.9} & \textbf{48.0} & \textbf{52.0} &
78.3 \downtinya{(↓24.5\%)} &
263.7 \downtinya{(↓45.3\%)}\\

\bottomrule

\end{tabular}
\end{adjustbox}
\vspace{-1mm}
\caption{\textbf{Comprehensive evaluation results on OVO-Bench across three categories:} 
\textbf{(i)} \textit{Real-Time Visual Perception} (OCR: Optical Character Recognition, ACR: Action Recognition, ATR: Attribute Recognition, STU: Spatial Understanding, FPD: Future Prediction, OJR: Object Recognition), 
\textbf{(ii)} \textit{Backward Tracing} (EPM: Episodic Memory, ASI: Action Sequence Identification, HLD: Hallucination Detection), 
\textbf{(iii)}\textit{Forward Active Responding} (REC: Repetition Event Count, SSR: Sequential Steps Recognition, CRR: Clues Reveal Responding). 
``ViT Enc. Latency'' is the time (s) required for ViT to encode 16 frames, and ``LLM Pref. Latency'' is the time (s) required for LLM prefilling.
}
\vspace{-4mm}
\label{tab:ovo-bench}
\end{table*}

\subsection{STC-Pruner: Dual-Anchor Pruning for LLM}

Despite an accelerated ViT by our STC-Cacher, the LLM prefilling stage remains a bottleneck due to the long context sequence $Z$ and the quadratic complexity of self-attention. Existing token pruning methods~\cite{liu2025vidcom2, tao2025dycoke, Zhang:SparseVLM} are often designed for offline, query-aware settings, making them unsuitable for streaming inference where decisions must be made without knowledge of the user query or future frames. This leads to our second research question:

\vspace{0.3em}
\textit{\textbf{In a query-agnostic and future-agnostic streaming context, what principle can serve as a reliable proxy for token importance to guide token pruning?}.}
\vspace{0.3em}

To address this, we introduce \textbf{STC-Pruner} (Figure~\ref{fig:STC-Pruner}), a \textbf{dual-anchor pruning} strategy to reduce context length for efficient LLM prefilling. STC-Pruner prunes tokens redundant to both the accumulated history and the current frame's context. The process is below:

\noindent \textbf{(I) Anchor Establishment:}  
We establish two anchors to model the past and present context in streaming videos, guiding the whole token pruning process:

\begin{itemize}
    \item \textbf{Temporal Context Anchor (TCA):} The mean of the historical buffer, representing the aggregated context of the recent past in streaming, is \( a_{\text{temporal}} = \frac{1}{|\mathcal{H}|} \sum_{h \in \mathcal{H}} h \), where \( \mathcal{H} = \{h_1, ..., h_{t-1}\} \) is the buffer containing the mean token vectors of the past \( W \) frames.
    
    \item \textbf{Spatial Context Anchor (SCA):} The mean of the token set from the \textit{current} frame, representing its global context or background information, is given by \( a_{\text{spatial}} = \frac{1}{N} \sum_{i=1}^N z_i \), where \( Z = \{z_i\}_{i=1}^N \) represents the sequence of \( N \) visual tokens for the current frame.
\end{itemize}

\noindent \textbf{(II) Dynamics Scoring:}  
Each token $z_j$ is scored based on its joint dissimilarity to both anchors. Using cosine distance ($d_{\text{cos}}(u, v) = 1 - \text{sim}(u, v)$), the dynamics score $S(z_j)$ is the product of its distances to the temporal and spatial anchors. This multiplicative formulation prioritizes tokens that are distinct from \textit{both} the past and the current frame:
\begin{equation} S(z_j) = \alpha \cdot d_{\text{cos}}(z_j, a_{\text{temporal}}) + (1-\alpha) \cdot d_{\text{cos}}(z_j, a_{\text{spatial}}) \end{equation}

where $d_{\text{cos}}$ is the cosine distance between the token $z_j$ and the anchors $a_{\text{temporal}}$ and $a_{\text{spatial}}$.

\noindent \textbf{(III) Token Pruning:}  
Given a \textit{pruning ratio} $R_\text{Pruner}$, we retain the top-$k$ tokens with the highest novelty scores, where $k = \lfloor N \cdot (1 - R_\text{Pruner}) \rfloor$ is determined by the preset pruning ratio $R_\text{Pruner}$. For example, with $R_\text{Pruner} = 25\%$, we prune 75\% of the tokens, retaining the top 25\% based on their dynamics scores. The pruned token set $Z' = \{z_j | z_j \in \text{TopK}(Z, k, \text{key}=S)\}$ is then passed to the LLM. After processing, the current frame's spatial context anchor $a_{\text{spatial}}$ is appended to the history buffer $\mathcal{H}$ (and the oldest entry is removed) to update the context for the next time step.

The token pruning process is formulated as:
\begin{equation}
Z' \leftarrow \text{TopK}(Z, k, \text{key}=S)
\end{equation}
where the Top-$k$ selection retains the tokens with the highest novelty scores, based on the joint dissimilarity to both the temporal and spatial anchors.

In this way, STC-Pruner reduces the context length by pruning redundant tokens while preserving important temporal and spatial information. By using dual anchors, STC-Pruner ensures that only the most relevant tokens are retained, enabling efficient LLM prefilling and reducing the computational burden in streaming video understanding.

\section{Experiments}

\subsection{Experimental Setup}


\begin{table*}[t]
\centering
\small
\setlength\tabcolsep{8pt}
\begin{adjustbox}{max width=\textwidth}
\begin{tabular}{lcccccccccc|c|cc}
\toprule
\makecell[l]{\textbf{Methods}} & \makecell[c]{\textbf{CS}} & \makecell[c]{\textbf{OP}} & \makecell[c]{\textbf{ATP}} & \makecell[c]{\textbf{PR}} & \makecell[c]{\textbf{ACP}} & \makecell[c]{\textbf{SU}} & \makecell[c]{\textbf{EU}} & \makecell[c]{\textbf{CT}} & \makecell[c]{\textbf{TR}} & \makecell[c]{\textbf{CR}}

& \makecell[c]{\textbf{Overall}} &
\makecell[c]{\textbf{\shortstack{ViT Enc. \\ Latency}}} & 

\makecell[c]{\textbf{\shortstack{LLM Pref. \\ Latency}}} \\
\midrule
\multicolumn{14}{c}{\textbf{End-to-End Online VideoLLMs with STC-Cacher}} \\
\midrule
\textcolor[rgb]{ .502,  .502,  .502}{StreamForest\tiny\texttt{(NIPS25)}} & \textcolor[rgb]{ .502,  .502,  .502}{82.7} & \textcolor[rgb]{ .502,  .502,  .502}{83.1} & \textcolor[rgb]{ .502,  .502,  .502}{84.3} & \textcolor[rgb]{ .502,  .502,  .502}{76.9} & \textcolor[rgb]{ .502,  .502,  .502}{75.6} & \textcolor[rgb]{ .502,  .502,  .502}{69.1} & \textcolor[rgb]{ .502,  .502,  .502}{77.5} & \textcolor[rgb]{ .502,  .502,  .502}{54.4} & \textcolor[rgb]{ .502,  .502,  .502}{78.2} & \textcolor[rgb]{ .502,  .502,  .502}{82.8} & \textcolor[rgb]{ .502,  .502,  .502}{77.3} & \textcolor[rgb]{ .502,  .502,  .502}{103.7} & \textcolor[rgb]{ .502,  .502,  .502}{366.2} \\
\rowcolor[HTML]{DAEFF9}
\textbf{+STC-Cacher} & 82.3 &81.7 & 81.6 & 78.7 & 75.6 & 67.9 & 78.8 & 62.2 & 75.1 & 81.3 & 76.9 & 67.7\downtinya{(↓34.7\%)} & 366.2 \\

\midrule
\multicolumn{14}{c}{\textbf{Offline-to-Online Framework ReKV with Token Compression Methods}} \\
\midrule

\textcolor[rgb]{ .502,  .502,  .502}{ReKV\tiny\texttt{(ICLR25)}} & \textcolor[rgb]{ .502,  .502,  .502}{79.2} & \textcolor[rgb]{ .502,  .502,  .502}{77.5} & \textcolor[rgb]{ .502,  .502,  .502}{75.6} & \textcolor[rgb]{ .502,  .502,  .502}{66.0} & \textcolor[rgb]{ .502,  .502,  .502}{62.2} & \textcolor[rgb]{ .502,  .502,  .502}{60.3} & \textcolor[rgb]{ .502,  .502,  .502}{72.3} & \textcolor[rgb]{ .502,  .502,  .502}{43.6} & \textcolor[rgb]{ .502,  .502,  .502}{69.7} & \textcolor[rgb]{ .502,  .502,  .502}{79.5} & \textcolor[rgb]{ .502,  .502,  .502}{69.1} & \textcolor[rgb]{ .502,  .502,  .502}{103.7} & \textcolor[rgb]{ .502,  .502,  .502}{482.4} \\

+ToMe\tiny\texttt{(ICLR23)} & 67.5 & 64.6 & 66.3 & 63.0 & 58.4 & \underline{53.3} & \textbf{65.2} & 19.7 & 57.3 & \underline{76.6} & 59.4 & 70.5\downtinya{(↓32.0\%)} & 257.8\downtinya{(↓46.6\%)} \\

+VisionZip\tiny\texttt{(CVPR25)} & 69.5 & 66.4 & 69.3 & 50.7 & 52.9 & \textbf{57.2} & \underline{64.1} & 33.3 & 54.8 & 73.4 & 60.4 & 103.7 & 258.3\downtinya{(↓46.5\%)} \\

+VidCom$^2$\tiny\texttt{(EMNLP25)} & \textbf{76.0} & \underline{68.1} & \underline{71.6} & 62.0 & \underline{58.6} & 52.0 & 64.0 & 42.5 & 60.4 & \underline{76.6} & 63.6 & 103.7 & 259.1\downtinya{(↓46.3\%)} \\

\rowcolor[HTML]{DAEFF9}
\textbf{+STC-Pruner}& \underline{75.4} & 66.8 & 71.2 & \underline{63.9} & 57.8 & 51.2 & 64.0 & \textbf{45.1} & \underline{63.2} & \underline{76.6} & \underline{63.7} & 103.7 & 259.2\downtinya{(↓46.3\%)} \\

\rowcolor[HTML]{A8D7F4}
\textbf{+STC-Cacher \& Pruner} & 74.4 & \textbf{71.3} & \textbf{72.9} & \textbf{69.4} & \textbf{58.9} & 52.4 & 60.9 & \underline{44.0} & \textbf{66.9} & \textbf{77.3} & \textbf{65.2} & 78.3\downtinya{(↓24.5\%)} & 263.7\downtinya{(↓45.3\%)} \\
\bottomrule
\end{tabular}
\end{adjustbox}
\vspace{-3mm}
\caption{\textbf{Comprehensive evaluation results on StreamingBench across real-time understanding tasks:} 
\textbf{(i)} \textit{Real-Time Visual Understanding} (OP: Object Perception, CR: Causal Reasoning, CS: Clips Summarization, ATP: Attribute Perception, EU: Event Understanding, TR: Text-Rich Understanding, PR: Prospective Reasoning, SU: Spatial Understanding, ACP: Action Perception, CT: Counting).}
\vspace{-2mm}
\label{tab:streamingbench}
\end{table*}


\begin{table*}[t]
\small
\centering
\setlength{\abovecaptionskip}{0.2cm}
\begin{adjustbox}{max width=\textwidth}
\setlength\tabcolsep{12pt} 

\begin{tabular}{lccccccc}
\toprule 
\multirow{2}{*}{\textbf{Methods}} &
\multirow{2}{*}{\textbf{EgoSchema}} &
\multirow{2}{*}{\textbf{MLVU-dev}} &
\multicolumn{4}{c}{\textbf{VideoMME}} &
\multirow{2}{*}{\textbf{Average}}
\\
\cmidrule(lr){4-7}
& & & \textbf{Short} & \textbf{Medium} & \textbf{Long} & \textbf{Overall}
& 
\\


\midrule 
\multicolumn{8}{c}{\textbf{End-to-End Online VideoLLMs with STC-Cacher}} \\
\midrule 
\textcolor[rgb]{0.502,0.502,0.502}{StreamForest{\tiny\texttt{(NIPS25)}}} & \textcolor[rgb]{0.502,0.502,0.502}{60.7} & \textcolor[rgb]{0.502,0.502,0.502}{68.9} & \textcolor[rgb]{0.502,0.502,0.502}{76.0} & \textcolor[rgb]{0.502,0.502,0.502}{58.6} & \textcolor[rgb]{0.502,0.502,0.502}{50.6} & \textcolor[rgb]{0.502,0.502,0.502}{61.7} & \textcolor[rgb]{0.502,0.502,0.502}{63.8} \\
\rowcolor[HTML]{DAEFF9}
\textbf{+STC-Cacher} & 59.7 & 68.1 & 73.0 & 57.4 & \textbf{51.8} & 60.7 & 62.8 \\

\midrule

\multicolumn{8}{c}{\textbf{Offline-to-Online Framework ReKV with Token Compression Methods}} \\
\midrule
\textcolor[rgb]{0.502,0.502,0.502}{ReKV{\tiny\texttt{(ICLR25)}}} & \textcolor[rgb]{0.502,0.502,0.502}{57.7} & \textcolor[rgb]{0.502,0.502,0.502}{68.6} & \textcolor[rgb]{0.502,0.502,0.502}{70.4} & \textcolor[rgb]{0.502,0.502,0.502}{55.4} & \textcolor[rgb]{0.502,0.502,0.502}{47.3} & \textcolor[rgb]{0.502,0.502,0.502}{57.7} & \textcolor[rgb]{0.502,0.502,0.502}{61.3} \\
+ToMe{\tiny\texttt{(ICLR23)}} & 55.2 & 63.1 & 59.6 & 52.0 & 43.4 & 51.7 & 56.7 \\
+VisionZip{\tiny\texttt{(CVPR25)}} & 55.8 & 63.2 & 59.6 & 51.8 & 43.4 & 51.6 & 56.9 \\
+VidCom$^2${\tiny\texttt{(EMNLP25)}} & \underline{60.6} & \underline{67.1} & \underline{68.2} & \underline{55.7} & \underline{46.6} & \underline{56.8} & \underline{61.5} \\
\rowcolor[HTML]{DAEFF9}
\textbf{+STC-Pruner} & \textbf{60.8} & \textbf{67.6} & \textbf{68.7} & \textbf{56.3} & 46.3 & \textbf{57.1} & \textbf{61.8} \\

\rowcolor[HTML]{A8D7F4}
\textbf{+STC-Cacher \& Pruner} & 59.0 & 67.0 & 67.3 & 53.9 & \textbf{48.3} & 56.5 & 60.8 \\
\bottomrule
\end{tabular}

\end{adjustbox}
\vspace{-1mm}
\caption{\textbf{Comprehensive evaluation results on three offline long video understanding benchmarks.}}
\vspace{-5mm}
\label{tab:offline-longvideo-bench}
\end{table*}

\noindent \textbf{Benchmarks.} We evaluate the proposed STC framework across two benchmark categories: \textbf{(i)} \textit{Streaming video understanding:} We use OVO-Bench~\cite{niu2025ovo-bench} and StreamingBench~\cite{lin2024streamingbench} to assess performance in streaming scenarios. \textbf{(ii)} \textit{Long video understanding:} We select EgoSchema~\cite{mangalam2023egoschema}, MLVU-dev~\cite{zhou2024mlvu}, and VideoMME~\cite{fu2025videomme} to evaluate effectiveness in offline long-video understanding.

\noindent \textbf{Implementations.} We evaluate two baseline VideoLLMs: \textbf{(i)} \textit{End-to-End Online Models:} We choose Dispider~\cite{qian2025dispider}, LiveCC~\cite{chen2025livecc}, and StreamForest~\cite{zeng2025streamforest}, which integrate sequence compression during training. We further accelerate their ViT encoding with STC-Cacher for efficient online inference. \textbf{(ii)} \textit{Offline-to-Online Frameworks:} We select ReKV~\cite{rekv}, with LLaVA-OV-7B~\cite{li2024llava-ov} as the backbone, to evaluate the impact of both STC-Cacher and STC-Pruner on ViT encoding and LLM prefilling efficiency, respectively. All models follow the original settings (0.5 fps protocol).

\noindent \textbf{Baselines.} We compare our STC with token compression methods like ToMe~\cite{Bolya:ToMe} for ViT encoding, and VisionZip~\cite{Yang2024:Visionzip} and VidCom$^2$~\cite{liu2025vidcom2} for LLM prefill efficiency. For STC-Cacher only, we set $\mathcal{N} = 4$ and $R_\text{Cacher} = 75\%$ to accelerate ViT encoding. For STC-Pruner and other methods reducing sequence length, we set $R_\text{Pruner} = 75\%$ to compress the sequence to 25\%. For STC-Cacher and STC-Pruner together (STC-Cacher \& Pruner), we set $\mathcal{N} = 2$ and $R_\text{Cacher} = 75\%$, and $R_\text{Pruner} = 70\%$ to compress the sequence to 30\%.

\subsection{Main Comparisons}


Table~\ref{tab:ovo-bench},~\ref{tab:streamingbench},~\ref{tab:offline-longvideo-bench} present comprehensive comparisons of our STC framework against other compression methods in streaming and long video understanding scenarios, revealing \textbf{\textit{three key advantages}} of our STC framework:

\noindent \textbf{(i) Outstanding Performance across Scenarios:} Our STC framework outperforms existing methods on all benchmarks. In Table~\ref{tab:ovo-bench} and~\ref{tab:streamingbench}, compared to the previous state-of-the-art method VidCom$^2$~\cite{liu2025vidcom2}, STC improves performance by \textbf{1.6} and \textbf{1.6} on OVO-Bench~\cite{niu2025ovo} and StreamingBench~\cite{lin2024streamingbench}, respectively, showcasing the strong performance of STC in streaming scenarios. Additionally, as shown in Table~\ref{tab:offline-longvideo-bench}, STC-Pruner achieves state-of-the-art results across three long video understanding benchmarks, demonstrating its robust performance across diverse scenarios.

\noindent \textbf{(ii) Consistent Efficiency Gains:} Both STC-Cacher and STC-Pruner are able to significantly improve ViT encoding and LLM prefilling efficiency. Notably, while both STC-Cacher and ToMe enhance ViT encoding efficiency, our method outperforms ToMe by \textbf{5.6}, \textbf{5.8}, and an average of \textbf{4.1} across OVO-Bench, StreamingBench, and three offline long video understanding benchmarks, demonstrating optimal performance-efficiency trade-offs. Furthermore, STC reduces ViT encoding latency and LLM prefilling latency by \textbf{24.5\%} and \textbf{45.3\%}, respectively, for ReKV on OVO-Bench.

\noindent \textbf{(iii) Model Compatibility:} Thanks to the two complementary compressors STC-Cacher and STC-Pruner, it jointly optimizes ViT encoding and LLM prefilling stages and is plug-and-play for end-to-end online models like Dispider~\cite{qian2025dispider}, LiveCC~\cite{chen2025livecc}, and StreamForest~\cite{zeng2025streamforest}, accelerating ViT encoding. Additionally, it speeds up both ViT encoding and LLM prefilling stages for offline-to-online framework ReKV~\cite{rekv}.

\begin{figure*}[t]
    \centering
    \includegraphics[width=\linewidth]{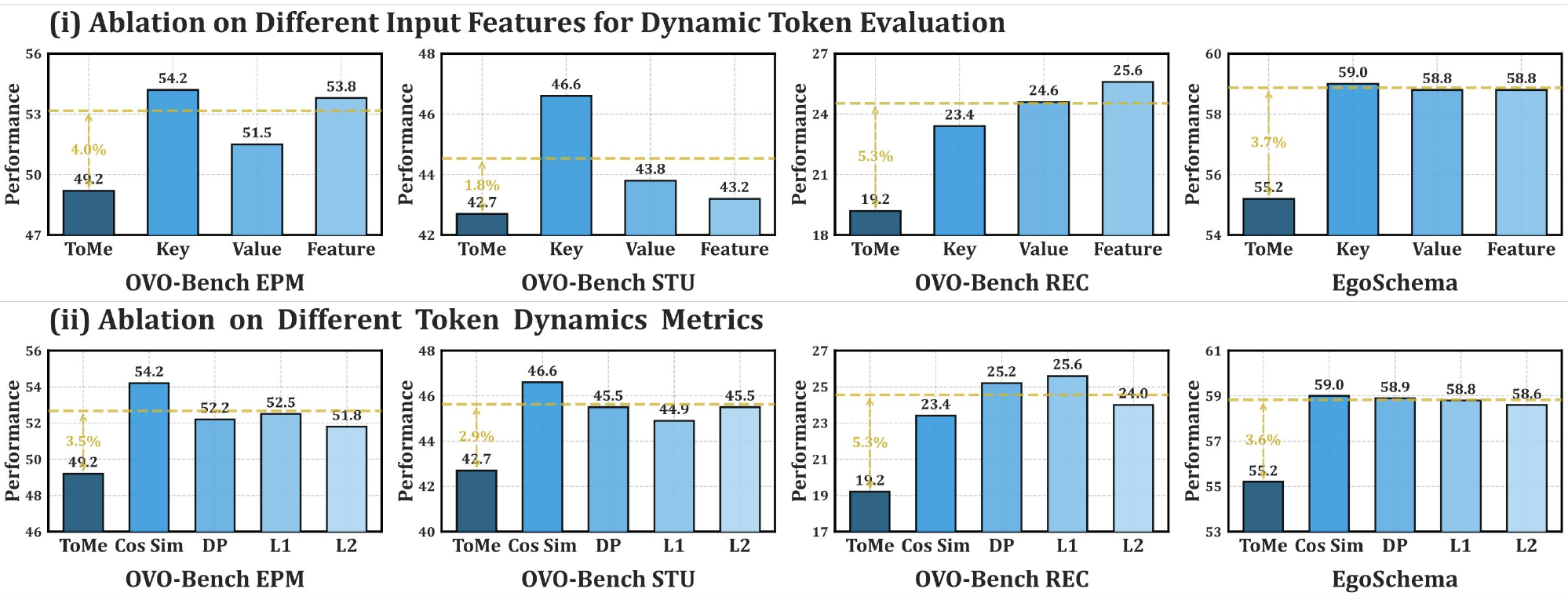}
    \vspace{-8mm}
    \caption{\noindent \textbf{Effects of Different Token Evaluation Strategies in STC-Cacher.} \textbf{(i)} Compares various features for dynamic token evaluation to identify the optimal dynamic token set for feature caching and reuse. \textbf{(ii)} Further compares different metrics for token dynamics evaluation, with "Cos Sim" referring to cosine similarity (smaller values indicate higher dynamics), "L1" and "L2" representing L1 and L2 distances (smaller values indicate higher dynamics), and "DP" denoting the dot product (smaller values indicate higher dynamics), respectively. The yellow line represents the average performance gap with ToMe~\cite{Bolya:ToMe}, indicating that STC-Cacher significantly outperforms ToMe.}
    \vspace{-5mm}
    \label{fig:ablation_12}
\end{figure*}

\subsection{Ablation Studies and Analysis}

We conduct ablation studies on STC-Cacher and STC-Pruner across three subsets of OVOBench (EPM, STU, and REC) and an offline long-video benchmark EgoSchema. 


\begin{table}[t] 
\small
\centering
\setlength{\abovecaptionskip}{0.2cm}
\begin{adjustbox}{max width=\columnwidth} 
\setlength\tabcolsep{6pt} 

\begin{tabular}{lcccc} 
\toprule 
\multirow{2}{*}{\textbf{Methods}} &
\multicolumn{3}{c}{\textbf{OVO-Bench}} & 
\multirow{2}{*}{\textbf{EgoSchema}}
\\
\cmidrule(lr){2-4} 
& \textbf{EPM} & \textbf{STU} & \textbf{REC} & 
\\
\midrule 



Attn ($R_\text{Cacher} = 85\%$) & 2.7 & 2.3 & 5.3 & 26.2 \\
MLP ($R_\text{Cacher} = 85\%$) & 49.8 & 43.2 & 25.3 & 57.1 \\
\rowcolor[HTML]{DAEFF9}
\textbf{Attn and MLP ($R_\text{Cacher} = 75\%$)} & \textbf{54.2} & \textbf{46.6} & \textbf{23.4} & \textbf{59.0} \\

\bottomrule 
\end{tabular}
\end{adjustbox}

\vspace{-1mm}
\caption{\textbf{Effects of different reusing features in STC-Cacher.} ``Attn'' and ``MLP'' are reusing features from attention and MLP.}
\vspace{-5mm}
\label{tab:ablation-update} 

\end{table}

\noindent \textbf{Dynamic token evaluation in STC-Cacher.} Figure~\ref{fig:ablation_12} aims to explore optimal strategies for identifying dynamic tokens to determine which can be cached and reused in STC-Cacher.

\noindent \textbf{(i) Effects of Different Features for Dynamics Evaluation:} We evaluate various features for token dynamics to identify the optimal basis. Results show that using key-states yields the best performance, likely because they capture a token's most informative aspects, including its historical relevance and contribution to attention, making them reliable for dynamic token identification.

\noindent \textbf{(ii) Effects of Different Token Dynamics Metrics:} We compare several metrics for token dynamics in ViT. All metrics outperform ToMe, showing that STC-Cacher’s feature caching-reuse strategy accelerates more gently, significantly outperforming ToMe in both streaming and offline settings. Cosine similarity consistently yields the best performance, so it is the default metric in STC-Cacher.

\noindent \textbf{Reusing Features across ViT Blocks.}  
Table~\ref{tab:ablation-update} investigates which intermediate activations should be reused during selective recomputation. Reusing only attention or MLP activations separately is insufficient: attention-only reuse leads to a performance drop, while MLP-only reuse also underperforms the combined strategy. This suggests that selective recomputation must preserve both the positional/contextual information in attention pathways and the channel/semantic representations in MLP pathways. Our hybrid strategy, reusing both, achieves the right balance, keeping cached tokens informative for downstream reasoning.


\begin{table}[t] 
\small
\centering
\setlength{\abovecaptionskip}{0.2cm}
\begin{adjustbox}{max width=\columnwidth} 
\setlength\tabcolsep{8pt} 

\begin{tabular}{lcccc} 
\toprule 
\multirow{2}{*}{\textbf{Methods}} &
\multicolumn{3}{c}{\textbf{OVO-Bench}} & 
\multirow{2}{*}{\textbf{EgoSchema}}
\\
\cmidrule(lr){2-4} 
& \textbf{EPM} & \textbf{STU} & \textbf{REC} & 
\\
\midrule 



Only $a_{\text{spatial}}$     & 50.5 & 47.2 & 25.8 & 59.9 \\
Only $a_{\text{temporal}}$    & 51.5 & 47.8 & 24.1 & 59.8 \\ 
\rowcolor[HTML]{DAEFF9}
\textbf{$a_{\text{spatial}}$ and $a_{\text{temporal}}$}  & \textbf{51.2} & \textbf{48.9} & \textbf{25.9} & \textbf{59.9} \\


\bottomrule 
\end{tabular}
\end{adjustbox}
\vspace{-1mm}
\caption{\textbf{Effects on different dynamics scoring in STC-Pruner.} $a_{\text{spatial}}$ and $a_{\text{temporal}}$ are using dynamics scoring by SCA and TCA.}
\vspace{-5mm}
\label{tab:ablation-scores} 

\end{table}

\noindent \textbf{Dynamics Scoring for STC-Pruner.}  
Table~\ref{tab:ablation-scores} breaks down the compression criteria of the STC-Pruner scoring mechanism. Using either the SCA $a_{\text{spatial}}$ or TCA $a_{\text{temporal}}$ alone leads to unbalanced performance: the former misses inter-frame novelty, while the latter overlooks intra-frame redundancy. The joint scoring approach, accounting for deviation from both anchors, consistently yields the most robust results. This highlights the need for the pruner to address both intra-frame redundancy and inter-frame dynamics to retain tokens essential for complex reasoning.

\section{Conclusion}

In this work, we propose Streaming Token Compression (STC), a plug-and-play framework designed to optimize both ViT encoding and LLM pre-filling stages for real-time streaming video understanding. Through two complementary modules, \textbf{STC-Cacher} and \textbf{STC-Pruner}, STC reduces redundant computations in ViT encoding and compresses token sequences before they enter the LLM, addressing key inefficiencies in streaming video understanding. Our approach significantly enhances processing efficiency while maintaining high accuracy. STC can be easily integrated into existing streaming VideoLLMs without the need for retraining, providing a practical and scalable solution for real-time deployment in latency-sensitive applications.

{
    \small
    \bibliographystyle{ieeenat_fullname}
    \bibliography{main}
}
\appendix

\clearpage
\setcounter{page}{1}
\maketitlesupplementary

In this appendix, we begin by elaborating on the detailed experimental settings in Section~\ref{sec:exp_details}, which covers specific descriptions of the benchmarks, model architectures, and baseline methods employed. Subsequently, Section~\ref{sec:appendix/ablation_studies} presents additional ablation studies, including performance comparisons on various model backbones, the impact of cache update intervals, and hyperparameter sensitivity analyses. Section~\ref{sec:appendix/algorithm} and Section~\ref{sec:appendix/more_vis} present the peudocode of our STC framework and more visualizations by STC-Cacher.


\section{Detailed Experiment Settings}
\label{sec:exp_details}

\noindent \textbf{Benchmark Details.}
We evaluate our STC on various video understanding benchmarks, detailed as follows:

\begin{itemize}
    \item \textbf{OVO-Bench}~\cite{niu2025ovo} is a benchmark for evaluating the online video understanding capabilities of Video-LLMs, with a focus on temporal awareness. It includes 644 unique videos, ranging from several minutes to half an hour in length, and features 2,814 human-curated meta-annotations with precise timestamps. The tasks are structured into three categories: Backward Tracing, Real-Time Visual Perception, and Forward Active Responding.
    \item \textbf{StreamingBench}~\cite{lin2024streamingbench} is a comprehensive benchmark designed to evaluate the streaming video understanding capabilities of MLLMs, emphasizing the gap between offline processing and real-time human-like interaction. It consists of 900 videos covering diverse scenarios and 4,500 human-curated QA pairs, where queries are presented at specific timestamps to simulate continuous inputs. The benchmark evaluates 18 distinct tasks organized into three core categories: Real-Time Visual Understanding, Omni-Source Understanding, and Contextual Understanding.

    \item 
    
    \textbf{EgoSchema}~\cite{mangalam2023egoschema} is a large-scale benchmark for long-form, egocentric video question answering, containing over 250 hours of video footage. It is designed to evaluate complex causal reasoning from a first-person perspective by asking ``why'' a particular action was performed. The benchmark consists of multiple-choice questions that require models to infer the actor's intent.
    
    \item \textbf{VideoMME}~\cite{fu2025videomme} comprises 900 videos and 2,700 multiple-choice questions across six domains, with durations from 11 seconds to 1 hour, categorized into short, medium, and long subsets.
    
    \item \textbf{MLVU}~\cite{zhou2024mlvu} (Massive Long Video Understanding) is a benchmark for long-form video comprehension, featuring 4,000 videos that range from 30 minutes to over 2 hours, totaling 4,800 hours. It comprises 13 challenging tasks designed to test multimodal and long-context reasoning, including character-centric question answering, plot analysis, and multi-event retrieval.
\end{itemize}

\noindent \textbf{Model Details.} 
We evaluate our STC on multiple VideoLLMs, including \textit{End-to-End Online VideoLLMs} and \textit{Offline-to-Online Frameworks}, detailed as follows:

\textbf{End-to-End Online VideoLLMs:}
\begin{itemize}
    \item \textbf{LiveCC}~\cite{chen2025livecc} explores the use of large-scale, cost-effective automatic speech recognition (ASR) transcripts for training Video LLMs. It proposes a novel streaming training approach that densely interleaves ASR words with video frames according to their timestamps, enabling the model to learn fine-grained, temporally-aligned vision-language modeling. To support this method, the work introduces two new datasets: \mbox{Live-CC-5M} for pre-training and \mbox{Live-WhisperX-526K} for supervised fine-tuning. This approach results in strong general video question answering (QA) performance while also exhibiting a novel capability for real-time video commentary.

    \item \textbf{StreamForest}~\cite{zeng2025streamforest} is a novel architecture specifically designed for efficient online video understanding in streaming scenarios. Unlike conventional Video-LLMs that either process entire videos offline or apply aggressive compression losing spatiotemporal details, StreamForest continuously analyzes video streams at 1~fps through a dual-memory mechanism. Central to the design is the Persistent Event Memory Forest (PEMF), which adaptively organizes video frames into hierarchical event-level tree structures guided by three penalty functions (temporal distance, content similarity, and merge frequency), enabling persistent long-term memory under strict token budgets. Complementing this, a Fine-grained Spatiotemporal Window (FSTW) captures detailed short-term visual cues for enhanced real-time perception. The work also introduces OnlineIT, an instruction-tuning dataset tailored for streaming video tasks, and ODV-Bench, a benchmark for autonomous driving scenarios. 
    \item \textbf{Dispider}~\cite{qian2025dispider} is a novel framework that enables active real-time interaction with Video Large Language Models by disentangling perception, decision, and reaction into asynchronous modules. Unlike previous streaming methods that alternate between video processing and response generation, Dispider features a lightweight scene-based perception module that continuously monitors video streams and dynamically segments them into non-uniform clips based on visual boundaries. A real-time decision module evaluates whether to trigger responses using special tokens ({\tt \textbackslash<TODO\textbackslash>} and {\tt \textbackslash<ANS\textbackslash>}) and historical context, while an asynchronous reaction module generates detailed responses without interrupting ongoing video processing. This non-blocking architecture ensures timely, contextually accurate responses for long-duration videos while maintaining computational efficiency. Dispider significantly outperforms existing streaming models on StreamingBench and demonstrates strong performance on conventional long-video benchmarks.
\end{itemize}

\textbf{Offline-to-Online Frameworks:}
\begin{itemize}
    \item \textbf{ReKV}~\cite{di2025streaming} is a novel, training-free framework designed to enable existing Video Large Language Models (Video-LLMs) with efficient Streaming Video Question-Answering (StreamingVQA) capabilities. Unlike traditional VideoQA systems that must process an entire video before responding to a query, ReKV continuously analyzes video streams in a streaming manner, allowing for prompt responses. During the encoding phase, the framework employs a sliding-window attention mechanism to reduce computational overhead, while simultaneously storing processed video KV Cache in memory to prevent information loss. When a query is posed, ReKV utilizes a retrieval module to load only the query-relevant KV-Caches to serve as context, enabling efficient answer generation. This design decouples the video encoding and question-answering processes, significantly enhancing efficiency and responsiveness, particularly when handling long videos.

\end{itemize}

\noindent \textbf{Baseline Details.} We compare our STC with below dominant token compression methods: 

\begin{itemize}
    \item 
    \textbf{ToMe}~\cite{Bolya:ToMe} is a token reduction method that improves Vision Transformer (ViT) throughput by gradually merging similar tokens across transformer layers, rather than pruning them. Unlike pruning-based approaches, ToMe can be applied off-the-shelf without retraining while maintaining compatibility with batched inference.
    \item \textbf{VidCom$^2$}~\cite{liu2025vidcom2} proposes a plug-and-play inference acceleration framework centered on frame uniqueness. Unlike methods using uniform compression, it employs a two-stage strategy: first dynamically adjusting compression intensity based on the distinctiveness of each video frame, and then performing adaptive token compression. Notably, it addresses the implementation constraints of previous methods by maintaining full compatibility with efficient operators like Flash Attention~\cite{Dao2022:FlashAttention}.
    \item \textbf{VisionZip}~\cite{Yang2024:Visionzip} introduces a text-agnostic token reduction framework applied before the LLM input. It identifies informative ``dominant'' tokens based on the self-attention weights within the vision encoder and aggregates the remaining redundant tokens through similarity-based merging to preserve contextual details.

\end{itemize}

\section{Additional Ablation Studies}
\label{sec:appendix/ablation_studies}

In this section, we present further quantitative analysis of our proposed method.
\noindent \textbf{Results on other models.} We first present the performance comparison on additional model architectures in Table~\ref{tab:ablation_streamforest} and Table~\ref{tab:ablation_streamforest_feature}.

\begin{table}[t]
\small
\centering
\setlength{\abovecaptionskip}{0.2cm}
\begin{adjustbox}{max width=\columnwidth}
\setlength\tabcolsep{8pt}

\begin{tabular}{lcccc}
\toprule
\multirow{2}{*}{\textbf{Methods}} &
\multicolumn{4}{c}{\textbf{OVO-Bench}} \\
\cmidrule(lr){2-5} 
& \textbf{EPM} & \textbf{STU} & \textbf{REC} & \textbf{Avg} \\
\midrule

Attn ($R_\text{Cacher} = 85\%$) & 30.6 & 33.9 & 11.2 & 25.2 \\

MLP ($R_\text{Cacher} = 85\%$) & 56.9 & 43.3 & 27.5 & 42.6 \\

\rowcolor[HTML]{DAEFF9}
\textbf{Attn and MLP ($R_\text{Cacher} = 75\%$)} & \textbf{57.9} & \textbf{46.1} & \textbf{29.9} & \textbf{44.6} \\

\bottomrule
\end{tabular}
\end{adjustbox}
\vspace{-1mm}
\caption{\textbf{Effects of different reusing features in STC-Cacher with StreamForest.} ``Attn'' and ``MLP'' are reusing features from attention and MLP.}
\vspace{-2mm}
\label{tab:ablation_streamforest}
\end{table} 
\begin{table}[t]
\small
\centering
\setlength{\abovecaptionskip}{0.8cm}
\begin{adjustbox}{max width=\columnwidth}
\setlength\tabcolsep{14pt}

\begin{tabular}{lcccc}
\toprule
\multirow{2}{*}{\textbf{Methods}} &
\multicolumn{4}{c}{\textbf{OVO-Bench}} \\
\cmidrule(lr){2-5} 
& \textbf{EPM} & \textbf{STU} & \textbf{REC} & \textbf{Avg} \\
\midrule

Feature & 56.9 & 45.5 & 28.9 & 43.8 \\

Value & \textbf{58.3} & 44.9 & 29.8 & 44.3 \\
\rowcolor[HTML]{DAEFF9}
\textbf{Key} & 57.9 & \textbf{46.1} & \textbf{29.9} & \textbf{44.6} \\

\bottomrule
\end{tabular}
\end{adjustbox}
\vspace{-7mm}
\caption{\textbf{ Compares various features for dynamic token evaluation to identify the optimal dynamic token set for feature caching and reuse.} We compare the performance of reusing Feature, Value, and Key.}
\vspace{-2mm}
\label{tab:ablation_streamforest_feature}
\end{table} 

\begin{table}[t]
\small
\centering
\setlength{\abovecaptionskip}{0.2cm}
\begin{adjustbox}{max width=\columnwidth}
\setlength\tabcolsep{10pt}
\begin{tabular}{lcccc}
\toprule
\multirow{2}{*}{\textbf{Cache Interval}} &
\multicolumn{4}{c}{\textbf{OVO-Bench}} \\
\cmidrule(lr){2-5}
& \textbf{EPM} & \textbf{STU} & \textbf{REC} & \textbf{Avg} \\
\midrule
$\mathcal{N} = 1$ & 54.6 & 51.1 & 25.5 & 43.7 \\
$\mathcal{N} = 4$ & 52.2 & 42.7 & 24.9 & 39.9 \\
$\mathcal{N} = 7$ & 51.9 & 44.4 & 23.8 & 40.0 \\
$\mathcal{N} = 10$ & 50.8 & 46.1 & 22.9 & 39.9 \\
$\mathcal{N} = \infty$ & 44.1 & 38.2 & 18.3 & 33.5 \\
\bottomrule
\end{tabular}
\end{adjustbox}
\vspace{-1mm}
\caption{\textbf{Ablation study on the cache update interval $\mathcal{N}$.} We compare different update frequencies ranging from frame-by-frame updates ($\mathcal{N}=1$) to no updates ($\mathcal{N}=\infty$). Results indicate that more frequent updates consistently yield better performance.}
\vspace{-5mm}
\label{tab:ablation-interval}
\end{table}

\subsection{Impact of Cache Update Interval}

Table~\ref{tab:ablation-interval} investigates sensitivity to cache update interval $\mathcal{N}$. We vary $\mathcal{N}$ from frame-by-frame to static. Results show that \textbf{more frequent updates yield better performance}. Performance drops sharply as the interval becomes static, revealing feature drift. This confirms periodic updates are essential for tracking temporal dynamics.

\subsection{Hyperparameter Sensitivity Analysis}

\subsubsection{Token Retention Ratios}
Table~\ref{tab:ablation_ratio} analyzes the impact of token retention ratios in the Cacher and Pruner modules.

\begin{table}[t]
\small
\centering
\setlength{\abovecaptionskip}{0.2cm}
\begin{adjustbox}{max width=\columnwidth}
\setlength\tabcolsep{10pt} 
\begin{tabular}{lcccc}
\toprule
\multirow{2}{*}{\textbf{Ablation Settings}} &
\multicolumn{4}{c}{\textbf{OVO-Bench}} \\
\cmidrule(lr){2-5}
& \textbf{EPM} & \textbf{STU} & \textbf{REC} & \textbf{Avg} \\
\midrule

\multicolumn{5}{l}{\textbf{(a) Impact of Cacher Ratio}} \\
\midrule
$R_\text{Cacher} = 85\%$ & 54.9 & 45.5 & 24.5 & 41.6 \\
$R_\text{Cacher} = 75\%$ & 52.2 & 42.7 & 24.9 & 39.9 \\
$R_\text{Cacher} = 50\%$ & 50.8 & 46.6 & 25.4 & 40.9 \\

\midrule

\multicolumn{5}{l}{\textbf{(b) Impact of Joint Ratios}} \\
\midrule
$R_\text{Cacher} = 50\%$, $R_\text{Pruner} = 75\%$ & 51.5 & 46.1 & 25.6 & 41.1 \\

$R_\text{Cacher} = 75\%$, $R_\text{Pruner} = 50\%$ & 50.8 & 46.2 & 26.9 & 41.3 \\

$R_\text{Cacher} = 75\%$, $R_\text{Pruner} = 75\%$ & 50.8 & 46.1 & 24.6 & 40.5 \\

\bottomrule
\end{tabular}
\end{adjustbox}
\vspace{-1mm}
\caption{\textbf{Ablation studies on Cacher and Pruner hyperparameters.} We analyze (a) the impact of the Cacher update ratio $R_\text{Cacher}$, and (b) the joint effect of $R_\text{Cacher}$ and the Prun ratio $R_\text{Pruner}$.}
\label{tab:ablation_ratio}
\end{table}

\begin{itemize}
    \item \textbf{Impact of Cacher Ratio ($R_{\text{Cacher}}$):} As shown in Table~\ref{tab:ablation_ratio} (a), a higher update ratio generally yields superior performance, as refreshing more tokens preserves dynamic information against rapid changes.
    \item \textbf{Impact of Joint Ratios:} Table~\ref{tab:ablation_ratio} (b) shows the interplay between Cacher and Pruner ratios. Performance remains robust across combinations, demonstrating that our framework does not rely on narrow hyperparameter tuning.
\end{itemize}

\begin{table}[t]
\small
\centering
\setlength{\abovecaptionskip}{0.2cm}
\begin{adjustbox}{max width=\columnwidth}
\setlength\tabcolsep{10pt}
\begin{tabular}{lcccc}
\toprule
\multirow{2}{*}{\textbf{Metrics}} &
\multicolumn{4}{c}{\textbf{OVO-Bench}} \\
\cmidrule(lr){2-5}
& \textbf{EPM} & \textbf{STU} & \textbf{REC} & \textbf{Avg} \\
\midrule
$a_{\text{temporal}}$ +$a_{\text{spatial}}$  & 51.2 & 48.9 & 25.9 & 42.0 \\
$a_{\text{temporal}}$ +2$a_{\text{spatial}}$  & 51.1 & 48.8 & 26.0 & 42.0 \\

2$a_{\text{temporal}}$ +$a_{\text{spatial}}$ & 50.1 & 47.7 & 25.9 & 41.6 \\

\bottomrule
\end{tabular}
\end{adjustbox}
\vspace{-1mm}
\caption{\textbf{Effects of balancing hyper-parameter $\alpha$ between $a_{\text{temporal}}$ and $a_{\text{spatial}}$.}}
\label{tab:ablation-pruner-balance}
\end{table}

\subsubsection{Spatial-Temporal Balance}
Finally, we investigate the hyperparameters used to combine spatial ($a_{\text{spatial}}$) and temporal ($a_{\text{temporal}}$) scores in Table~\ref{tab:ablation-pruner-balance}. Results show that performance is optimal when scores are balanced or slightly favor the spatial term. Heavily up-weighting the temporal score leads to degradation, implying that while temporal dynamics are useful, spatial semantics remain the fundamental basis for reasoning.

\section{Algorithm Pseudocode}
\label{sec:appendix/algorithm}

In this section, we provide the detailed pseudocode for the two stages of our Streaming Token Compression (STC) framework. Algorithm~\ref{alg:stc_cacher} details the cache-aware selective computation in the ViT (STC-Cacher), and Algorithm~\ref{alg:stc_pruner} describes the dual-anchor pruning mechanism for the LLM input (STC-Pruner).

\section{More Visualizations by STC-Cacher}
\label{sec:appendix/more_vis}

\begin{figure*}[!t]
  \centering
   \includegraphics[width=\linewidth]{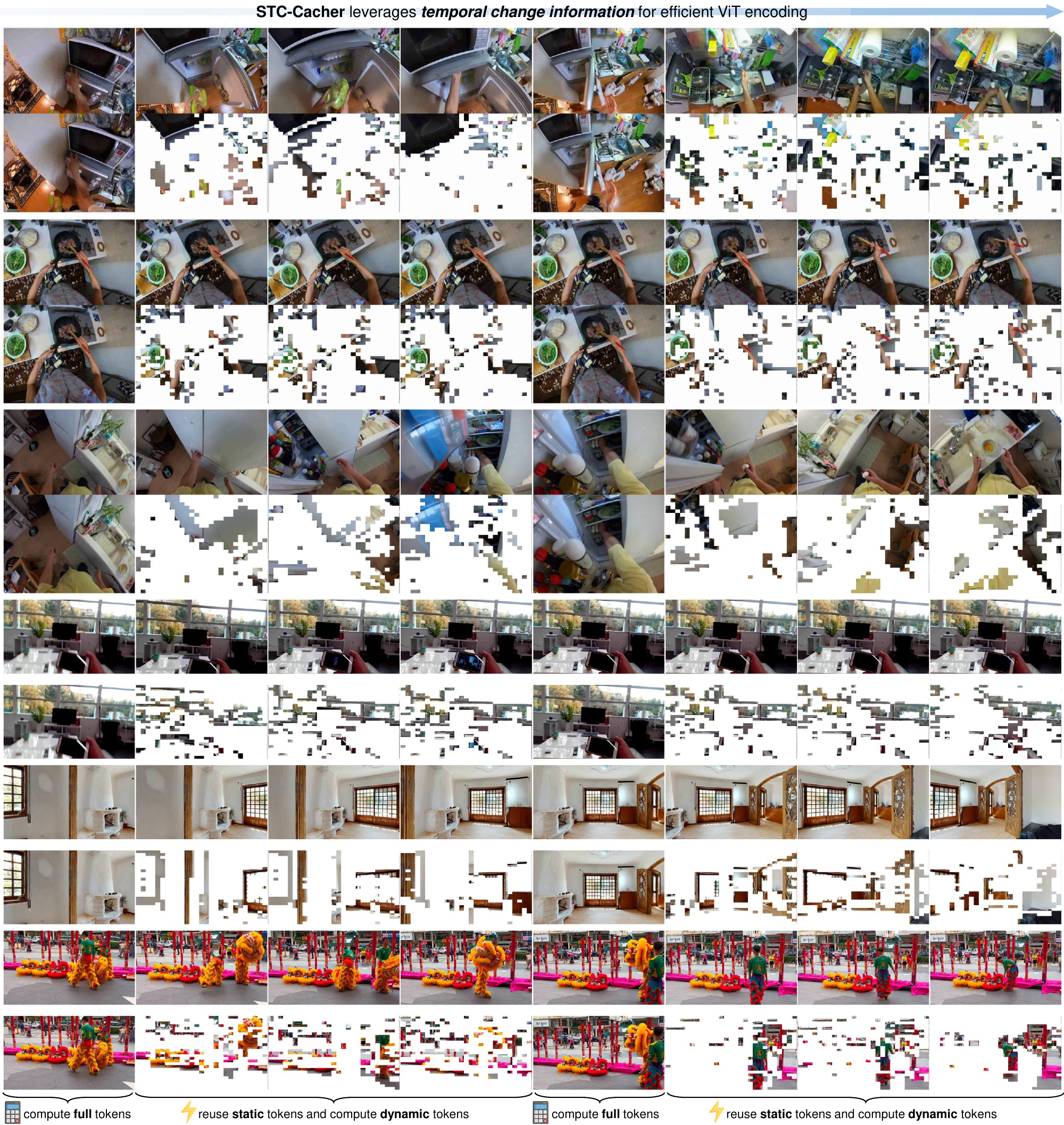}
   \vspace{-7mm}
    \caption{\textbf{More visualization of cache-aware selective computation by STC-Cacher.}}   
   \label{fig:more_caching_visualizations}
   \vspace{-5mm}
\end{figure*}

Figure~\ref{fig:more_caching_visualizations} presents more visualization results by STC-Cacher, which intuitively demonstrates that STC-Cacher can concentrate computational overhead on temporally significant regions while compressing computation for visual tokens that remain static across temporal dimensions.

\begin{algorithm}[h]
\caption{STC-Cacher (ViT Acceleration)}
\label{alg:stc_cacher}
\begin{algorithmic}[1]
\REQUIRE Video frame sequence $\mathbf{V} = \{\mathbf{v}_t\}_{t=1}^T$, Cache Interval $\mathcal{N}$, Cache Reuse Ratio $R_{\text{Cacher}} \in [0, 1)$
\ENSURE Sequence of visual tokens $\mathbb{Z} = \{\mathbf{Z}_t\}_{t=1}^T$

\STATE Initialize output list $\mathbb{Z} \leftarrow \emptyset$

\FOR{$t = 1 \to T$}
    \IF{$t \pmod \mathcal{N} == 1$ \OR $t == 1$}
        \STATE \textit{// (I) Reference Frame: Full Computation}
        \STATE Perform full forward pass on $\mathbf{v}_t$
        \STATE Cache representations: $\mathcal{C}_{\text{ref}}^l \leftarrow \{ \mathbf{K}_{\text{ref}}^l, \mathbf{V}_{\text{ref}}^l, \mathbf{A}_{\text{ref}}^l, \mathbf{M}_{\text{ref}}^l \}$ for all layers $l$
        \STATE $\mathbf{Z}_t \leftarrow$ Output from final layer
    \ELSE
        \STATE \textit{// (II) Non-reference Frame: Selective Computation}
        \FOR{each layer $l$}
            \STATE \textit{// 1. Identify Dynamic Tokens}
            \STATE $S_{\mathbf{f}} \leftarrow \text{CosSim}(\mathbf{K}_{\text{curr}}^l, \mathbf{K}_{\text{ref}}^l)$
            \STATE $\mathcal{I}_{\mathbf{f}} \leftarrow \underset{i}{\text{arg top-k}} (1 - S_{\mathbf{f}}[i])$ \quad where $k = \lfloor N_{tok} \cdot (1 - R_{\text{Cacher}}) \rfloor$
            
            \STATE \textit{// 2. Selective Attention with Scatter Update}
            \STATE Compute Query/Value for $\mathcal{I}_{\mathbf{f}}$: $Q_{\text{sel}}^l, V_{\text{sel}}^l$
            \STATE $\bar{V}_{\mathbf{f}}^l \leftarrow \mathbf{V}_{\text{ref}}^l$; \quad $\bar{V}_{\mathbf{f}}^l[\mathcal{I}_{\mathbf{f}}] \leftarrow V_{\text{sel}}^l$
            \STATE $A_{\text{sel}}^l \leftarrow \text{Attention}(Q_{\text{sel}}^l, \bar{V}_{\mathbf{f}}^l)$
            \STATE $\bar{A}_{\mathbf{f}}^l \leftarrow \mathbf{A}_{\text{ref}}^l$; \quad $\bar{A}_{\mathbf{f}}^l[\mathcal{I}_{\mathbf{f}}] \leftarrow A_{\text{sel}}^l$ 

            \STATE \textit{// 3. Selective MLP with Scatter Update}
            \STATE $M_{\text{sel}}^l \leftarrow \text{MLP}(\text{LN}(\bar{A}_{\mathbf{f}}^l[\mathcal{I}_{\mathbf{f}}]))$ 
            \STATE $\bar{M}_{\mathbf{f}}^l \leftarrow \mathbf{M}_{\text{ref}}^l$; \quad $\bar{M}_{\mathbf{f}}^l[\mathcal{I}_{\mathbf{f}}] \leftarrow M_{\text{sel}}^l$
        \ENDFOR
        \STATE $\mathbf{Z}_t \leftarrow$ Output from final layer $\bar{M}_{\mathbf{f}}^L$
    \ENDIF
    \STATE Append $\mathbf{Z}_t$ to $\mathbb{Z}$
\ENDFOR
\RETURN $\mathbb{Z}$
\end{algorithmic}
\end{algorithm}
\begin{algorithm}[t]  
\caption{STC-Pruner (LLM Input Compression)}
\label{alg:stc_pruner}
\begin{algorithmic}[1]
\REQUIRE Visual tokens $\mathbf{Z}_t$ (from STC-Cacher), History Buffer $\mathcal{H}$, Pruning Ratio $R_{\text{Pruner}} \in [0, 1)$, Balance Factor $\alpha$
\ENSURE Pruned tokens $\mathbf{Z}'_t$, Updated Buffer $\mathcal{H}$

\STATE \textit{// (I) Anchor Establishment}
\STATE $a_{\text{temporal}} \leftarrow \frac{1}{|\mathcal{H}|} \sum_{h \in \mathcal{H}} h$ \quad \textit{(Historical Context)}
\STATE $a_{\text{spatial}} \leftarrow \frac{1}{|{\mathbf{Z}_t}|} \sum_{z \in \mathbf{Z}_t} z$ \quad \textit{(Current Frame Context)}

\STATE \textit{// (II) Dynamics Scoring}
\FOR{each token $z_j \in \mathbf{Z}_t$}
    \STATE Calculate joint dissimilarity:
    \STATE $S(z_j) \leftarrow \alpha \cdot d_{\text{cos}}(z_j, a_{\text{temporal}}) + (1-\alpha) \cdot d_{\text{cos}}(z_j, a_{\text{spatial}})$
\ENDFOR

\STATE \textit{// (III) Token Pruning}
\STATE Calculate retention count: $k' \leftarrow \lfloor |\mathbf{Z}_t| \cdot (1 - R_{\text{Pruner}}) \rfloor$
\STATE $\mathbf{Z}'_t \leftarrow \text{TopK}(\mathbf{Z}_t, k', \text{key}=S)$

\STATE \textit{// (IV) Context Update}
\STATE $\mathcal{H} \leftarrow \text{Enqueue}(\mathcal{H}, a_{\text{spatial}})$ \quad \textit{(Update history for next step)}

\RETURN $\mathbf{Z}'_t$
\end{algorithmic}
\end{algorithm} 


\end{document}